\documentclass[12pt]{article}
\usepackage[margin=1in]{geometry}
\usepackage{amsmath}
\usepackage{pslatex}
\usepackage{pdflscape}
\usepackage[colorlinks,urlcolor=blue,citecolor=blue,linkcolor=blue]{hyperref}
\usepackage{setspace}
\usepackage{authblk}
\usepackage{lineno} 

\usepackage{fontspec} 
\usepackage[normalem]{ulem} 
\usepackage{xcolor}

\usepackage{array,booktabs, longtable, tabularx, adjustbox} 

\usepackage{tablefootnote}

\usepackage{graphicx} 
\usepackage{caption}
\usepackage{csquotes}
\usepackage[style=apa, backend=biber, sorting=nyt,uniquename=false]{biblatex}
\usepackage{gb4e} 
\noautomath 

\newcommand{\revise}[1]{\textcolor{black}{#1}}
\title{Graded strength of comparative illusions is explained by Bayesian inference}
\author[1,2]{Yuhan Zhang}
\author[1]{Erxiao Wang}
\author[1]{Cory Shain}
\affil[1]{Department of Linguistics, Stanford University \\

Margaret Jacks Hall, Building 460 Rm. 127, Stanford, CA 94305-2150}
\affil[2]{Corresponding author, yuhanczhang@gmail.com}
\date{}

\begin{document}

\maketitle


\begin{abstract}

Like visual processing, language processing is susceptible to illusions in which people systematically misperceive stimuli. In one such case --- the comparative illusion (CI), e.g., \textit{More students have been to Russia than I have} --- comprehenders tend to judge the sentence as acceptable despite its underlying nonsensical comparison. Prior research has argued that this phenomenon can be explained as Bayesian inference over a noisy channel: the posterior probability of an interpretation of a sentence is proportional to both the prior probability of that interpretation and the likelihood of corruption into the observed (CI) sentence. Initial behavioral work has supported this claim by evaluating a narrow set of alternative interpretations of CI sentences and showing that comprehenders favor interpretations that are more likely to have been corrupted into the illusory sentence. In this study, we replicate and go substantially beyond this earlier work by directly predicting the strength of illusion with a quantitative model of the posterior probability of plausible interpretations, which we derive through a novel synthesis of statistical language models with human behavioral data. Our model explains not only the fine gradations in the strength of CI effects, but also a previously unexplained effect caused by pronominal vs.\ full noun phrase \textit{than}-clause subjects. These findings support a noisy-channel theory of sentence comprehension by demonstrating that the theory makes novel predictions about the comparative illusion that bear out empirically. This outcome joins related evidence of noisy channel processing in both illusory and non-illusory contexts to support noisy channel inference as a unified computational-level theory of diverse language processing phenomena.

\textbf{Keywords:} 
linguistic illusion, the noisy-channel theory, Bayesian inference, sentence acceptability judgments, language comprehension
\end{abstract}

\section{Introduction}

Illusions offer a window into the mind: when our perceptions deviate from reality, they reveal the processes by which we make meaning from a complex and ambiguous environment. An important insight in recent decades is that a surprisingly diverse range of illusions are consistent with \textit{Bayesian inference}, that is, with a perceiver that does not merely process sensory information bottom-up, but instead integrates sensory information with top-down prior expectations to arrive at an optimal percept \parencite{von-Helmholtz1924-el,Gershman2015-jf,Griffiths2001-rr,Yang2021-mv,Born2020-ic,Geisler2002-xd,Weiss2002-lx,Nour2015-wt}. In other words, rather than being failures of perception, illusions may demonstrate why perception is so reliably successful: reliance on prior knowledge can compensate for the noise and ambiguity that frequently characterize our sensory environment. From this perspective, illusions provide powerful convergent support for Bayesian theories of cognition more generally \parencite{Tenenbaum2011-ev,Pouget2013-se,Gershman2019-wa}.

Although illusion research tends to focus on core senses like sight \parencite{Weiss2002-lx}, hearing \parencite{Goh2023-uq}, smell \parencite{Herz2001-vw}, and touch \parencite{de-Vignemont2005-uw}, it has been known for decades that a higher-order mental faculty --- language --- is also vulnerable to illusions \parencite{gibson1999memory,leivada2020acceptable, phillips2011grammatical,parker2025illusion}. \revise{A language illusion refers to a phenomenon where a linguistic stimulus (e.g., a sentence) systematically misleads comprehenders into accepting a grammatical error, a semantic anomaly, or a misinterpretation.} A well-known case is the semantically anomalous \textit{comparative illusion} (CI), which has been attested in real language use \parencite{montalbetti1984after} and is exemplified in (\ref{ex:ci-sentence}).
\begin{exe} 
    \ex \textbf{Comparative illusion:} More people have been to Russia than I have. \label{ex:ci-sentence}
\end{exe}

\begin{exe} 
    \ex \textbf{Control:} More people have been to Russia than I have been to Russia. \label{ex:ci-sentence-expand}
\end{exe}

The (typically undetected) anomaly in (\ref{ex:ci-sentence}) is that there is no degree or quantity provided by the \textit{than} clause that can be plausibly compared with the quantity of people in the main clause, making the whole sentence nonsensical. The anomaly becomes obvious when the elided verb phrase (\textit{been to Russia}) is made explicit as in (\ref{ex:ci-sentence-expand}). Comprehenders overwhelmingly accept (\ref{ex:ci-sentence}) at first sight and only realize after careful consideration that they are unsure of its meaning \parencite{liberman2018, pullum2004, pullum2009}.

In striking parallel to research on visual illusions, language illusions have historically been explained as the result of \revise{errorful heuristic processing \parencite{christianson_when_2016,ferreira2003misinterpretation, ferreira2007good,paape_quadruplex_2020} or shallow processing \parencite{sanford2002depth}, but have recently been offered an alternative explanation via a general Bayesian theory of language comprehension}: \textit{noisy channel inference}, according to which comprehenders interpret language by relying on Bayes' rule to integrate their percept of the noisy linguistic signal with top-down prior expectations about plausible meanings that their interlocutor likely intends to convey \parencite{gibson_rational_2013,levy_noisy-channel_2008}. Intuitively, noisy channel theories predict that language illusions will occur when the anomalous input is similar in form and meaning to something that a speaker would be likely to say. Formally, noisy channel theories predict that illusions will occur when the \textit{posterior probability} $p(s_i|s_p)$ of an intended sentence $s_i$ given a perceived (illusory) sentence $s_p$ is high:
\begin{equation} \label{eq:bayes}
    p(s_i|s_p) = \frac{p(s_i)p(s_p|s_i)}{p(s_p)} \propto p(s_i)p(s_p|s_i)
\end{equation}
As shown in Eq.\ref{eq:bayes}, the posterior is proportional to both the prior probability over intended sentences (or meanings) $p(s_i)$ and the likelihood $p(s_p|s_i)$ of $s_i$ being corrupted into $s_p$ by transmission over a noisy channel that is susceptible to production, perception, and environmental noise.
This view has recently been shown to both accommodate and make novel successful predictions about a growing range of language illusion types \parencite{hahn2022resource,qian2023comprehenders,ryskin_agreement_2021,zhang_noisy-channel_2023,zhang_comparative_2024}.

Recently, \textcite{zhang_comparative_2024} offered empirical support for a noisy-channel account of the comparative illusion by showing that participants' interpretational preferences for CI sentences followed their estimated likelihood for the plausible interpretations to be corrupted into the CI sentence during production, consistent with the predicted proportional relationship between posterior probability $p(s_i|s_p)$ and noise likelihood $p(s_p|s_i)$. \revise{However, this study had several limitations. First, they did not investigate how prior probabilities of interpretations, $p(s_i)$, affect inference, thus failing to model the full open-ended posterior probability.
Second, they only focused on explaining the most illusory CI sentence such as (\ref{ex:ci-sentence}) and left other types of CI sentences in the literature unexplained. Those CI types differ from (\ref{ex:ci-sentence}) by having noun-phrase \textit{than}-clause subjects and showed gradations in illusion strength that no prior theory predicts \parencite{oconnor2013,oconnor2015comparative,wellwood2018anatomy}. In this study, we both replicate key findings of \textcite{zhang_comparative_2024} at scale and go substantially beyond them by (i) defining a quantitative approximation of the CI posterior distribution, representing priors and noise likelihoods by probabilities derived from statistical language models \parencite{radford2019language, zhang2022opt} and independently collected human behavioral data, and (ii) using the approximation to predict variability in the illusion strength of multiple types of comparative illusions.
Results show that our quantitative model of noisy channel theory makes graded predictions about comparative illusion strength that track human acceptability judgments, supporting the hypothesis that these illusion effects reflect (roughly) Bayes-optimal sentence interpretation.}

Our key contributions are as follows. \textit{First}, we replicate (under a nearly order-of-magnitude increase in sample size) the key finding from \textcite{zhang_comparative_2024} that illusion effects in CI are reflected in sentence acceptability ratings: CI sentences are rated equally acceptable to felicitous controls and far better than non-illusory semantic anomalies. \textit{Second}, we show that fine gradations in illusion strength between CI sentences \revise{caused by morphosyntactic differences in the \textit{than}-clause subjects} are predicted by our quantitative model of the posterior probability, confirming a novel prediction made by the noisy-channel theory. \textit{Third}, we show that our quantitative estimate of the posterior explains variation in acceptability over and above strong controls, supporting noisy-channel Bayesian inference as a theory of sentence comprehension with broad empirical coverage, including the comparative illusion. This outcome converges with other evidence \parencite[e.g.,][]{lau2017grammaticality, keller2000gradience, hofmeister2013source} to suggest that acceptability ratings provide a rich window into the processes of sentence comprehension that goes well beyond the original intent of quantifying grammaticality.

\section{Background}
\label{sect:background}

Since the discovery of the comparative illusion \parencite{montalbetti1984after}, most accounts have invoked ``mental edits'' that comprehenders might apply to derive a plausible interpretation from the CI sentence. \revise{According to the analysis-by-synthesis model in \textcite{townsend2001sentence}, sentence (\ref{ex:ci-sentence}) could be an integration of two frequent and felicitous sentence templates (\ref{ex:townsend_sentence}) with high string overlap (underlined), where segments outside of the parentheses are ``blended'' into the canonical CI sentence. Comprehenders initially accept the perceived sentence because of the associated plausible templates but soon realize that it is incoherent after careful syntactic analysis. This model predicts a two-stage comprehension process where comprehenders would eventually realize the anomaly after an initial acceptance. Additional mechanism needs to be added to explain why the illusion lasts during untimed acceptability tasks \parencite{zhang_comparative_2024} and what underlying interpretation comprehenders hold about the anomalous sentence that makes them insensitive to the anomaly.}
\begin{exe}
    \ex \begin{xlist} \label{ex:townsend_sentence}
        \ex More \underline{people have been to Russia than I} (could believe). 
        \ex \underline{People have been to Russia} (more) \underline{than I} have.
    \end{xlist}
\end{exe}
%

Others have proposed subtler ``edits'' that could transform the illusory sentence into a well-formed one. The range of possibilities is exemplified in (\ref{ex:repair}):
\begin{exe}
    \ex \label{ex:repair}
    \begin{xlist}
        \ex \textbf{Comparative Illusion}: More judges vacationed in Florida than the lawyers did. \label{ex:edit-illusion}
        \ex \textbf{Event comparison (shift \textit{more})}: Judges vacationed in Florida \underline{more} than the lawyers did. \label{ex:wellwood-ev-comp}
        \ex \textbf{Individual comparison (delete \textit{the})}: More judges vacationed in Florida than \sout{the} lawyers did. \label{ex:individual1}
        \ex \textbf{Individual comparison (insert \textit{of the})}: More \underline{of the} judges vacationed in Florida than the lawyers did.  \label{ex:individual2}
    \end{xlist}
\end{exe}
\textcite{oconnor2015comparative} and \textcite{wellwood2018anatomy} proposed that the sentence-initial \textit{more} in CI sentences, e.g., (\ref{ex:edit-illusion}), could be moved to an adverbial position, e.g., (\ref{ex:wellwood-ev-comp}). The edited sentence compares the frequencies with which the two relevant parties (judges and lawyers) engaged in the activity (vacationing in Florida). For ease of reference, we label this interpretation ``event comparison'', following \textcite{wellwood2018anatomy}. \textcite{oconnor2015comparative} did not test this hypothesis empirically, and a subsequent study \parencite{wellwood2018anatomy} failed to corroborate its predictions: participants in a recall task did not correct sentences such as (\ref{ex:edit-illusion}) into (\ref{ex:wellwood-ev-comp}).

Alternatively, the matrix or the \textit{than}-clause noun phrase subjects could be modified instead \parencite{oconnor2013,oconnor2015comparative}. The edited sentences, e.g., (\ref{ex:individual1}) and (\ref{ex:individual2}), compare the cardinality of the two sets (the number of judges who vacationed in Florida and the number of lawyers who vacationed in Florida). We label this interpretation ``individual comparison''. Even though O'Connor and colleagues did not test this hypothesis experimentally, subsequent work has provided empirical support \parencite{de2016more,kelley2018more}. 

Finally, \textcite{christensen2016dead} offered a third possible interpretation based on evidence from Danish: \textit{(Some) people have been to Russia, except for me}. Since this interpretation negates participation in the event by the \textit{than}-clause subject, we label it ``event negation''. Due to morphosyntactic differences between English \textit{more} and Danish \textit{flere} (where \textit{flere} can only be used as the comparative morpheme for the nominal modifier \textit{mere} ``many'', not the adverbial modifier \textit{meget} ``much''), it is not guaranteed that this ``event negation'' interpretation generalizes to English.

Although these heuristic accounts of CI shed light on the alternative interpretations of CI, they do not explain why those interpretations are considered and to what extent they are preferred over others during comprehension. \revise{
\textcite{paape2024linguistic} was the first to propose that two general psycholinguistic theories of nonliteral sentence interpretation might fill this explanatory gap for CI: noisy-channel theories (reviewed above) and ``good-enough'' processing theories \parencite{ferreira2002good,ferreira2007good,christianson_when_2016} according to which comprehenders minimize the effort they expend subject to their current communicative goals, potentially leading to errorful interpretations.
\textcite{paape2024linguistic} pitted these theories against each other under the assumption that noisy-channel inference requires detection (i.e., conscious awareness) of errors in order to correct them, whereas good-enough processing does not.
Paape showed that the majority of participants in a metalinguistic judgment task coded CI sentences as ``Get it'' and a subset of them additionally coded them as ``Incorrect'', which was taken to indicate that participants sometimes (but not always) consciously recognized the anomaly and inferred a (non-veridical) plausible interpretation, consistent with noisy-channel predictions.
This mixed result was taken to support a contribution of both kinds of processing (noisy-channel and good-enough) to the comparative illusion.
However, in its standard formulation \parencite{levy_noisy-channel_2008,gibson_rational_2013}, noisy-channel theory is a computational-level \parencite{marr2010vision} theory that makes the fewest possible commitments about the mechanisms of error correction, and thus takes no stance on whether these mechanisms require conscious awareness.
Under this interpretation of the two theories, noisy-channel inference and good-enough processing are not distinguishable mental processes, but rather (potentially compatible) theories stated at different levels of analysis (respectively, computational and algorithmic). 
This computational-level interpretation of noisy-channel theory is the one that we will adopt in this work, setting aside the finding of variation in the awareness of the anomaly reported by \textcite{paape2024linguistic}, which we nonetheless consider an important explanandum for psycholinguistics generally.}

\revise{To our knowledge, the first study to develop and test a mathematically explicit noisy-channel theory of CI effects was \textcite{zhang_comparative_2024}, who showed that the noisy-channel model accurately predicted human preference rankings for CI interpretations.}
Under their account, comprehenders probabilistically resolve the meaning conveyed by noisy linguistic signals according to Bayes' theorem (Eq. \ref{eq:bayes}, repeated below for convenience):
\begin{equation} \tag{\ref{eq:bayes}}
    p(s_i|s_p) = \frac{p(s_i)p(s_p|s_i)}{p(s_p)} \propto p(s_i)p(s_p|s_i)
\end{equation}
They treated CI sentences as the perceived sentence $s_p$ and the interpretations as $s_i$. Imagine someone intends to convey message  $s_i$ but errorfully produces the anomalous CI sentence $s_p$ (which may be similar in form or meaning to $s_i$). To infer the producer's intended meaning, the comprehender considers the posterior probability $p(s_i | s_p)$ of possible intended messages $s_i$ given $s_p$, which is proportional to the prior probability $p(s_i)$ that someone intends to convey that meaning, times the noise likelihood $p(s_p|s_i)$, i.e., the probability that $s_i$ will be corrupted into $s_p$ in the noisy channel.
For a given intended message, the more probable it is for someone to convey that message and/or the more likely for corruptions and distortions to transform the message into the perceived signal, the more probable that intended message is to be chosen as the intended interpretation \parencite[][a.o.]{gibson_rational_2013,liu2020structural,poliak2024not,poliak2025,zhang_noisy-channel_2023}.

The key empirical support for this view in \textcite{zhang_comparative_2024} was as follows. Via an acceptability judgment task (Exp. 1), they showed that the canonical CI sentence (\ref{ex:ci-sentence}) elicited a strong illusion effect: illusory sentences had almost identical acceptability ratings to the felicitous controls (e.g., \textit{Many people have been to Russia more than I have}). Via a paraphrase task where participants were asked to provide the meaning behind CI sentences (Exp. 2), they identified three common interpretations in (\ref{ex:zhang-3interpretations}), supporting the claims and intuitions from \textcite{oconnor2013}, \textcite{oconnor2015comparative}, \textcite{christensen2016dead}, and \textcite{wellwood2018anatomy}. 
\begin{exe}
    \ex \label{ex:zhang-3interpretations}
    \begin{xlist}
    \ex \textbf{Event comparison (shift \textit{more}):} People have been to Russia \underline{more} than I have. \label{ex:zhang-eventcomp}
    \ex \textbf{Individual comparison (\textit{than}-clause substitution):} More people have been to Russia than \underline{just me}. \label{ex:zhang-individual}
    \ex \textbf{Event negation (negation-comparative substitution):} More people have been to Russia \underline{but I haven't}. \label{ex:zhang-negate}
    \end{xlist}
\end{exe}
Via a forced-choice task where participants chose the most probable interpretation of CI out of the three (Exp. 3), the ``event comparison'' interpretation (\ref{ex:zhang-eventcomp}) was chosen most frequently, followed by ``individual comparison'' (\ref{ex:zhang-individual}) and ``event negation'' (\ref{ex:zhang-negate}). This ranking of interpretation prominence was taken as a behavioral proxy for the posterior probability $p(s_i|s_p)$, which aligned with the ranking of a noise likelihood proxy $p(s_p|s_i)$ that was obtained by asking a separate group of participants to rate the likelihood for someone to utter the CI sentence when intending to convey one of the three plausible alternatives (Exp. 4). This positive correlation between proxies of $p(s_i|s_p)$ and $p(s_p|s_i)$ is consistent with noisy-channel predictions, assuming the differences of priors of these three alternatives do not systematically outweigh those of noise likelihood.

Nonetheless, both \textcite{paape2024linguistic} and \textcite{zhang_comparative_2024} stopped short of quantitatively modeling the full noisy-channel posterior probability distribution of plausible alternatives of a given CI sentence.
In addition, \textcite{zhang_comparative_2024} reported that CI sentences with pronominal \textit{than}-clause subjects (\ref{ex:pro-sing} \& \ref{ex:pro-plu}) have a stronger illusion effect -- comprehenders regard them as more acceptable -- compared to those with full noun phrase \textit{than}-clause subjects (\ref{ex:np-sing} \& \ref{ex:np-plu}). Without a quantitative model of the posterior, it remains unclear whether the noisy-channel theory can account for it.
\begin{exe} 
\ex \label{ex:ci-four-exs}
\begin{xlist}
\ex More students have been to Russia than \textbf{I have}. \label{ex:pro-sing}
\ex More students have been to Russia than \textbf{we have}. \label{ex:pro-plu}
\ex More students have been to Russia than \textbf{the teacher has}. \label{ex:np-sing}
\ex More students have been to Russia than \textbf{the teachers have}. \label{ex:np-plu}
\end{xlist}
\end{exe}

Our key contribution in this work is to bridge this gap by proposing a quantitative approximation of the posterior, leveraging the open-ended string probabilities provided by statistical language models (LMs) and integrating these with an open-ended behavioral measure of semantic interpretation.
\revise{We show that this posterior approximation explains unique variance in the acceptability ratings of the CI sentences, even with control variables that have been reported to affect sentence acceptability (e.g., sentence probability, order of presentation in an experiment)}. The statistical modeling work offers a strong piece of evidence supporting that noisy-channel inference underlies the processing of comparative illusion sentences.

\section{Data availability}
All the experimental materials, the collected data, and the analysis scripts can be accessed via the Open Science Framework repository at \href{https://osf.io/exn5k/?view_only=7abfb773cf04438fb80a8eb388f94b5b}{https://osf.io/exn5k}.

\section{Experiment 1} \label{sec:exp1}

Experiment 1 was an acceptability judgment task of anomalous CI sentences. High acceptability ratings of CI sentences were assumed to be the behavioral signature of an illusion effect, which is the main observation we aim to explain with the noisy-channel theory. The results in Experiment 1 show this signature, and also reveal variable CI illusion strength, replicating \textcite{zhang_comparative_2024}. The acceptability ratings from Experiment 1 were then treated as the dependent variable in the statistical models reported in Section \ref{sec:intro_modeling_method}.

\subsection{Participants} 

500 participants were recruited from the crowd-sourcing platform Prolific.com, a tenfold increase in sample size over \textcite{zhang_comparative_2024}. Each participant was paid \$6 for their participation. We excluded data from those (a) who did not complete at least 90\% of all questions; (b) who did not answer at least 75\% of the comprehension checks correctly; (c) who gave the same acceptability rating or answer to comprehension questions across all trials; and (d) who self-identified as non-native speakers of English or not born in the United States. 494 participants contributed to the final analysis.

Both experiments were conducted in accordance with the IRB approved protocol at the affiliated institution. All participants provided their consent to the study before doing the tasks.

\subsection{Materials \& Procedure} \label{sec:exp1-mat}

The materials were identical to those from Experiment 1 in \textcite{zhang_comparative_2024}. Example (\ref{ex:exp1-material}) shows the condition manipulation. The illusory materials had a 2 $\times$ 2 within-subjects manipulation, crossing the form of the \textit{than}-clause subject (pronoun vs. noun phrase) and its number feature (singular vs. plural) (\ref{ex:exp1-material}). Two control conditions consisted of acceptable sentences. The control sentences for the pronoun condition (\ref{ex1:pro-control}) substituted the sentence-initial \textit{more} with \textit{many} and had \textit{more} as an adverb that modifies the matrix verb phrase. The control sentences for the NP condition had bare plural noun phrases as the \textit{than}-clause subject (\ref{ex1:np-control}), where the difference from the target illusory conditions lies in the (morpho)syntactic structure of the \textit{than}-clause subject. 

\begin{exe} 
\ex \label{ex:exp1-material}
\begin{xlist}
\ex More students have been to Russia than I have. (pronoun, singular, illusory) \label{ex1:pro-sing}
\ex More students have been to Russia than we have. (pronoun, plural, illusory)  \label{ex1:pro-plu}
\ex More students have been to Russia than the teacher has. (NP, singular, illusory) \label{ex1:np-sing}
\ex More students have been to Russia than the teachers have. (NP, plural, illusory) \label{ex1:np-plu}
\ex Many students have been to Russia more than I have. (pronoun, control, acceptable) \label{ex1:pro-control}
\ex More students have been to Russia than teachers have. (NP, control, acceptable) \label{ex1:np-control}
\end{xlist}
\end{exe}

There were 30 critical items in each of the six conditions and 64 plausible fillers. Participants read each critical item in one of the conditions as well as all fillers, for a total of 94 trials per participant. The critical trials were presented according to a Latin Square design: (i) each participant read the same number of critical trials for each condition and only read one trial for each item, and (ii) all items across all conditions received equal number of responses for the entire experiment. The conditions and presentation order were randomized by participant. See all data and code in the online repository (\href{https://osf.io/exn5k/?view_only=7abfb773cf04438fb80a8eb388f94b5b}{https://osf.io/exn5k}).

For each trial, participants first completed a YES/NO comprehension question (e.g., \textit{Does this sentence mention students and Russia?}) as an attention check and then gave their judgment rating according to the instruction prompt (i.e., \textit{How natural is the sentence above?}) on a fully-labeled seven-point Likert scale (1 = ``Extremely unnatural'', 2 = ``Unnatural'', 3 = ``Somewhat unnatural'', 4 = ``Neutral'', 5 = ``Somewhat natural'', 6 = ``Natural'', 7 = ``Extremely natural''). The correct answer to the comprehension question was counterbalanced such that half of the items were YES and half were NO. This design helped filter out inattentive participants who stuck with one response throughout the experiment. Two practice trials preceded the critical experiment. One trial featured a natural sentence and the other featured an unnatural sentence.

\subsection{Results}

\begin{figure}[!ht]
    \centering
    \includegraphics[width=0.8\linewidth]{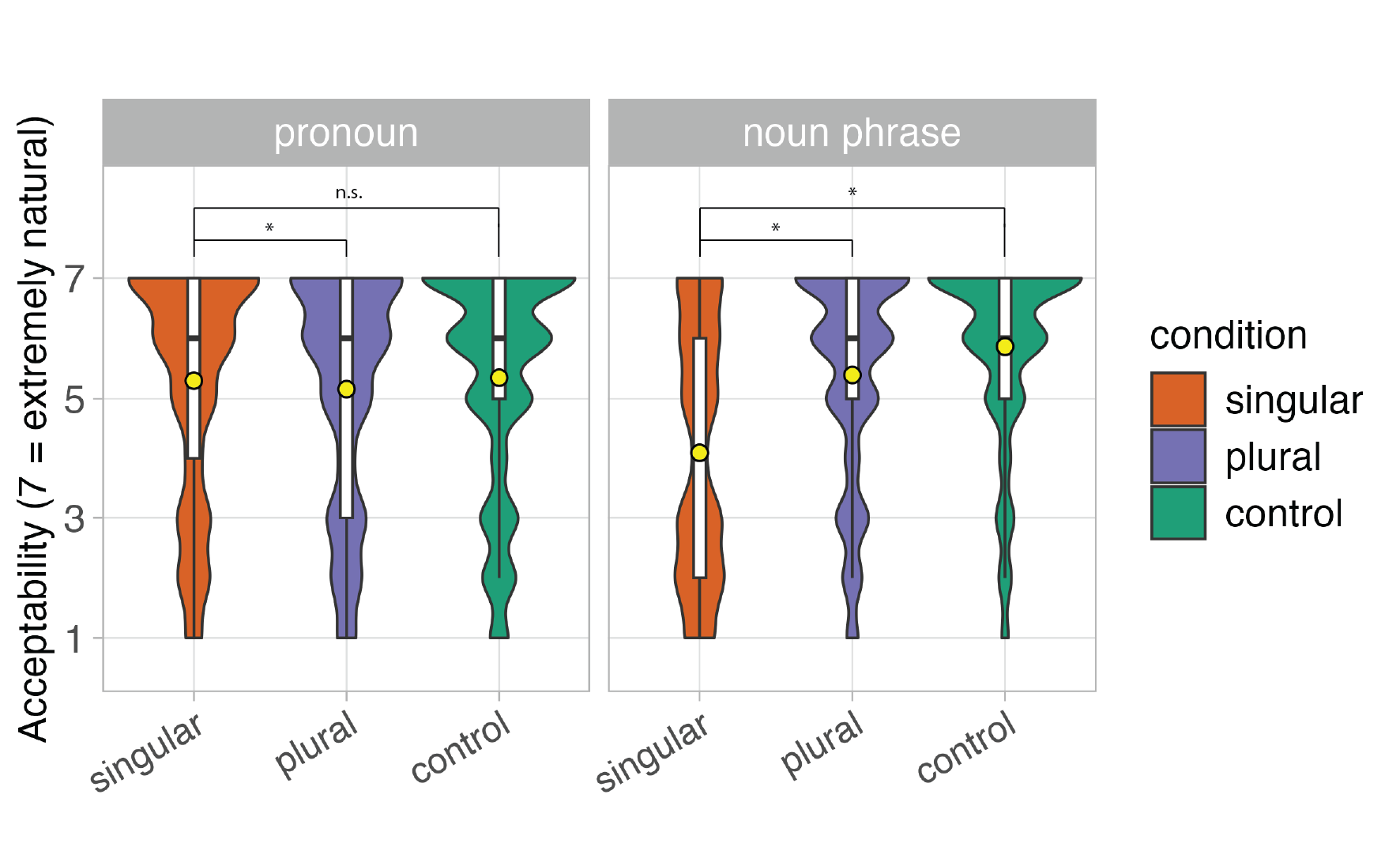}
    \caption{Acceptability ratings across the six conditions in Experiment 1 \newline (The violin plots represent the density of rating distributions with wider sections indicating higher concentration of data points. The white boxes represent the inter-quartile range and the short black horizontal line represents the median. The yellow dots represent the mean. * represents statistical significance and n.s. represents the opposite.)}
    \label{fig:exp1-acceptability}
\end{figure}

The participants who met inclusion criteria provided a combined total of 14,820 responses to critical items. Figure \ref{fig:exp1-acceptability} displays the acceptability ratings across the six conditions in (\ref{ex:exp1-material}). The similarity in acceptability ratings between each illusory condition and its acceptable control condition indicates the illusion strength. The singular pronoun condition is the most illusory (most similar in acceptability to its corresponding control), followed by the plural pronoun condition. The noun phrase conditions show a degradation of illusion strength, as evidenced by a decreasing acceptability from the plural condition to the singular condition. The pronoun control condition is rated less natural than its noun phrase counterpart \parencite[also replicating][]{zhang_comparative_2024}. This could be due to the increased length of the pronoun control sentences and/or the use of a lower-frequency syntactic structure \textit{many...more than}.  

To statistically investigate the effects of \textit{than}-clause subject form (pronoun vs.\ noun phrase) and number on CI acceptability, we ran a Bayesian multilevel cumulative ordinal regression model \parencite{burkner2019ordinal}, using the \textit{brms} package \parencite{burkner2017brms} in R \parencite{r2024r}. The response variable was the raw acceptability score, an ordinal variable where the intervals between boundaries were not assumed to be equal and the boundaries were modulated by population-level fixed effects and group-level random effects. The fixed effects included two dummy-coded categorical variables capturing the \textit{than}-clause subject form and number features (form: reference level = pronoun, number: reference level = singular). The control condition was treated as a level in the number variable. An interaction term between the two fixed effects was also entered. The group-level random effects included random intercepts and random slopes of the full population-level effects by both participant and item (i.e., the 30 items that occurred in the six conditions), obtaining the maximal random-effect structure \parencite{barr2013random}.\footnote{We chose the current model with the maximal random effects over a base model with only random intercepts by participant and item based on models' predictive abilities, as evidenced by Bayesian leave-one-out cross-validation (LOO-CV)\parencite{vehtari2017practical}. See Table \ref{tab:appendix-model-comparison-exp1} for details.}

Both the intercept and the population-level fixed effects were given a Gaussian prior centered on 0 with a standard deviation of 2 ($\text{Normal}(0,2)$), following \textcite{zhang_comparative_2024}. Following \textcite{nalborczyk2019introduction}, the by-participant and by-item varying intercepts were given a weakly informative Half-Cauchy prior ($\text{Cauchy}(0,10)$).
The correlation matrices for the two sets of random effects were each assigned a prior of $\text{LKJ}(2)$. The remaining model parameters took default settings in \textit{brms}. The model had four sample chains each with 4000 iterations of the Markov Chain Monte Carlo algorithm. The first 1000 iterations were taken as warmup. The $\hat{R}$s for all fixed effects were 1.0, indicating successful convergence of the four sampling chains. We use $\beta$ to represent the mean of posterior estimates of predictor coefficients and $CrI$ to represent the 95\% credible interval. These coefficients are on the scale of the standard normal latent variable assumed in cumulative regression to underlie the observed values of the ordinal dependent variable.

The results show that, with $>$95\% probability, the pronoun plural condition is less acceptable than its singular counterpart ($\beta = -0.28, \text{CrI} = [-0.43, -0.14]$) and the pronoun singular condition does not differ from the pronoun control condition ($\beta = 0.02, \text{CrI} = [-0.31, 0.35]$).
In other words, singular pronoun CI sentences produce a strong illusion (their acceptability is statistically indistinguishable from controls), whereas plural pronoun CI sentences produce a slight but detectable drop in illusion strength. Because of the contrast coding (where the reference condition was pronoun singular), the model does not directly estimate posterior distributions over the control, singular, and plural conditions within the noun phrase conditions. Thus, we used non-linear hypothesis testing (via \texttt{hypothesis.brmsfit} in R, with Bonferroni-corrected $\alpha$ value) to inspect the illusion strengths in the noun phrase conditions. There is decreasing acceptability from control to singular conditions within the noun phrase conditions. The posterior probability for the noun phrase control condition to be more acceptable than the singular condition is nearly 1 with an effect size difference of 2.74 ($\text{CrI} = [2.31, 3.18]$). The posterior probability for the plural noun phrase condition to be acceptable than its singular counterpart is nearly 1, with an effect size difference of 1.86 ($\text{CrI} = [1.49, 2.21]$). The posterior probability for the noun phrase control condition to be more acceptable than the noun phrase plural condition is nearly 1, with an effect size difference of 0.88 ($\text{CrI} = [0.58, 1.2]$) (See the full results in Table \ref{tab:appendic-stat-exp1}).

\subsection{Discussion}

By examining the acceptability of various types of anomalous CI sentences in a large sample of participants (n=500), we confirm the gradience of CI effects caused by changes in the \textit{than}-clause subject form and number feature. While the general acceptability patterns align with \textcite{zhang_comparative_2024}, we additionally found that the pronoun plural condition is slightly less acceptable than its singular counterpart, an outcome which challenges the report in \textcite{zhang_comparative_2024} of no difference in acceptability. We attribute this disparity to the increased statistical power of the current study. We otherwise closely replicate \textcite{zhang_comparative_2024}, and turn now to \textit{explaining} this graded illusion strength.
To this end, we required a quantitative estimate of the noisy-channel posterior in Eq.~\ref{eq:bayes}.
To obtain such an estimate, we ran an additional experiment (Experiment 2).

\section{Experiment 2} \label{sec:exp2}

Experiment 2 aimed to approximate an empirical sample of $s_i$ given the corrupted CI sentence $s_p$.
The motivation for this experiment was two-fold: \textit{first}, the correction data are of direct interest due to the light they shed on human interpretive preferences for CI sentences, and \textit{second}, we sought to use these behavioral data to construct a model of the noisy-channel posterior $p(s_i | s_p)$ (section~\ref{sec:stats_model}). The experiment consisted of an open-ended behavioral task in which participants were instructed to make the fewest edits necessary to transform an illusory CI sentence into a plausible sentence and the corrected plausible sentence was treated as $s_i$. For each CI sentence $s_p$, Experiment 2 therefore provided a sample of plausible interpretations $s_i$.

\subsection{Participants}
207 participants were recruited from Prolific and each was paid \$5 for their participation. After going through the same screening method as in Experiment 1, 199 were included in the final analysis.

\subsection{Materials \& Design}

Participants were asked to act like a language editor, read a list of 30 unnatural English sentences, and make as few edits as possible to turn each sentence into a natural and meaningful one. They were instructed to type the complete edited sentence in a text box and avoid using artificial intelligence technologies (e.g., ChatGPT). \revise{This task is of course unlike the ordinary conditions of language use in many ways, and we do not claim that participants' edits are a direct window onto their subjective posterior probability $p(s_i | s_p)$.
We simply assume that their edits allow us to approximate this probability well enough for us to learn about the distribution of CI interpretations and test our noisy channel model quantitatively in the following section.
Given our focused interest in modeling the noisy-channel posterior, we are not seeking to study open-ended sentence editing as a process unto itself, just as a vehicle for estimating an implicit quantity that is impossible to measure directly (the subjective probabilities of intended sentences under the comprehender's assumed noise distribution).
Our instruction to participants to make as few edits as possible follows from this goal under the standard noisy-channel assumption that the noise distribution assigns higher probabilities to smaller corruptions (i.e., to perceived sentences that are more similar to the intended sentence).}

The unnatural sentences all came from the illusory conditions, e.g., (\ref{ex1:pro-sing})--(\ref{ex1:np-plu}), in Experiment 1. The 2 x 2 conditions were treated as a within-subjects manipulation. The 30 sentences appeared randomly according to a Latin Square design. \revise{Due to time and resource constraints on this relatively laborious editing task, no fillers or other types of anomalous sentences were included.} Before the sentence correction task, each trial had a YES/NO comprehension question as an attention check. At the beginning of the task, participants were reminded of how the word \textit{more} can be used in diverse linguistic environments. \textit{More} not only modifies the frequency of events, such as \textit{I play basketball more than tennis}, but also the number of individuals, such as \textit{More pumpkins are needed on Halloween than tomatoes}.

\subsection{Data Analysis}

We collected a total of 5,970 responses and adopted both quantitative and qualitative measures to analyze the corrected sentences. The quantitative measure characterizes the word-level edit distance of the two sentences which will be used to compute the noise likelihood term in the noisy-channel posterior model (Eq.\ref{eq:bayes}). The qualitative measure characterizes the meaning of the corrected sentence, highlighting how the syntactic distribution of \textit{more} in the corrected sentence determines the focus of comparison. 

\subsubsection*{Quantitative measures} \label{sec:exp2_edit_distance}

We calculated the word-level edit distance between a perceived sentence and a corrected sentence based on the Damerau-Levenshtein algorithm \parencite{damerau1964technique, levenshtein1966binary}.\footnote{The software package can be accessed through \hyperlink{https://github.com/lanl/pyxDamerauLevenshtein}{https://github.com/lanl/pyxDamerauLevenshtein}.} The Damerau-Levenshtein distance (DLD) between two strings is the minimum number of single word insertions, substitutions and deletions (and in some cases, word pair transpositions) necessary to derive one string from another. The DLD measure is taken as a proxy to represent the distance between the perceived and corrected sentences.

\subsubsection*{Qualitative measures}

We created a rule-based algorithm in Python\footnote{See the script in \href{https://osf.io/exn5k/overview?view_only=7abfb773cf04438fb80a8eb388f94b5b}{https://osf.io/exn5k/}} to detect critical syntactic patterns of word-level edits that are indicative of the possible meanings of the corrected sentences. The syntactic patterns are listed in (\ref{ex:exp2-detect-features}). The program integrated all syntactic patterns for a single correction according to rules in Table \ref{tab:ex2-feature-merge} (2nd column) and assigned an interpretation to the corrected sentence. The interpretation categories took inspiration from prior literature \parencite{oconnor2013,oconnor2015comparative,wellwood2018anatomy,zhang_comparative_2024} and the naturally occurring groups emerging from the data. After the Python program assigned an initial interpretation category, the first author manually checked and corrected false assignments. See Table \ref{tab:ex2-feature-merge} for the full range of interpretation categories, the rules of meaning assignment, and example sentences.\footnote{Among the meaning categories, although the ``incomplete comparison'' group qualifies as ``ungrammatical'', it occurred frequently enough in the experiment (n=136) that we created a separate label. Regarding the ``outlier'' group, we determined the outlier by first identifying item-wise mean and standard deviation of the word-level edit distance. Any trial of that item that deviated from the mean by three standard deviations was treated as an outlier.}

\begin{exe}
    \ex Critical syntactic patterns in the corrected sentences \label{ex:exp2-detect-features}
    \begin{xlist}
        \ex \textbf{Change of the syntactic environment of \textit{more}:} Possible patterns include ``transforming \textit{more} to other comparative structures such as \textit{compare}, \textit{less than}'' (e.g., \textit{I have vacationed \textbf{less} in Florida than most lawyers have}), ``shifting \textit{more} to non-sentence-initial positions'' (e.g., \textit{Unlike us, teenagers use TikTok \textbf{more}}), ``using \textit{more than} instead'' (e.g., \textit{Men have talked about the 2022 midterm elections \textbf{more than} the women have}), etc.
        
        \ex \textbf{Change of the \textit{than} clause:} Possible patterns include ``dropping the entire \textit{than} clause'' (e.g., \textit{We have more Americans who have enjoyed high school}), ``fronting the \textit{than} clause'' (e.g., \textit{More Canadians \textbf{than Americans} have toured Antelope Canyon}), etc.
        
        \ex \textbf{Change of the \textit{than}-clause subject:} Possible patterns include ``dropping the determiner \textit{the} in the NP condition'' (e.g., \textit{More Canadians than Americans have toured Antelope Canyon}), ``changing the singular NP to a plural one'' (e.g., ditto), ``changing the case of the pronoun subject'' (e.g., \textit{More adults have studied for their driver's license than \textbf{us}}), etc.
        
        \ex \textbf{Addition of negation:} Possible patterns include the insertion of negation-related words or tokens, such as \textit{not}, \textit{n't}, etc.
    \end{xlist}
\end{exe}

\revise{Two groups of corrections emerge from Table \ref{tab:ex2-feature-merge}: The first group indicates that participants might have had a clear and unambiguous understanding of one intended meaning of the perceived sentence and corrected it in a predictable way. This is supported by the homogeneous edit strategies across participants. We label this group as \textit{plausible} because the corrected sentences are plausible in meaning. This group included sentences labeled as ``event comparison'', ``individual comparison'', ``event negation'', and ``double comparison''. The second group indicates that participants might have lost track of the meaning or the syntactic representation of the sentence, or just been inattentive. Corrections in the second group (about 17\% of the data) are thus unpredictable, heterogeneous, and idiosyncratic across participants. The edited sentences are also problematic in nature. This \textit{implausible} group included categories of ``no change'', ``incomplete comparison'', ``blended'', ``outlier'', and ``ungrammatical''.}

\begin{longtable}{|>{\centering\arraybackslash}p{0.18\textwidth}|>{\raggedright\arraybackslash}p{0.3\textwidth}|>{\centering\arraybackslash}p{0.3\textwidth}|>{\centering\arraybackslash}p{0.15\textwidth}|}
  \caption{Summary of emergent interpretation categories of corrected sentences and rules of interpretation assignment in Experiment 2}\label{tab:ex2-feature-merge} \\ \hline
\textbf{Interpretation category} & \textbf{Interpretation assignment rule} & \textbf{Example} & \textbf{Plausibility} \\ \hline
\endfirsthead
    
\hline
Interpretation category & Interpretation assignment rule & Example & Plausibility \\ \hline
\endhead
    
\hline
\multicolumn{4}{|r|}{Continued on next page} \\ \hline
\endfoot
    
\hline
\endlastfoot

Event comparison & \textit{More} is used to modify the frequency of activities, involving the shift of \textit{more}, the appearance of \textit{more than}, or the transformation to other comparative structures such as \textit{less than}. & More men have talked about the 2022 midterm elections than the women have. $\rightarrow$ Men have talked about the 2022 midterm elections \textbf{more} than the women have. & Plausible \\ \hline
Individual comparison & \textit{More} is used to compare the number of individuals who have participated in the activity, usually involving fronting the \textit{than} clause, deleting the determiner \textit{the}, and pluralizing singular \textit{than}-clause subjects. & More white-collar workers have attempted the 10-minute workout than the blue-collar workers have. $\rightarrow$ More white-collar \textbf{than blue collar workers} have attempted the 10-minute workout. & Plausible \\  \hline
Event negation  & A negation word is added to deny the participation of the \textit{than}-clause subject. & More customers have thought about Black Friday shopping season in 2022 than I have. $\rightarrow$ More customers have thought about Black Friday shopping season in 2022, \textbf{but I am not one of them.} & Plausible \\ \hline
Double comparison & A second \textit{more} is inserted into the adverbial position of the matrix clause. & More novelists have loved alcohol than the movie stars have. $\rightarrow$ More novelists have loved alcohol \textbf{more} than the movie stars have. & Plausible\\ \hline
No change & The illusory sentence and the corrected one are exactly the same. & More Americans have toured Antelope Canyon than we have. $\rightarrow$ More Americans have toured Antelope Canyon than we have. & Implausible \\ \hline
Incomplete comparison & The original structure of \textit{more...than} disappears in the response, usually involving dropping \textit{more} and/or the \textit{than} clause. & More California residents have used UPS service than we have. $\rightarrow$ California residents have used UPS service than we have. & Implausible \\ \hline
Blended & The response simultaneously asserts both an ``event comparison'' reading and an ``individual comparison'' interpretation. & More students have been to Russia than I have. $\rightarrow$ Most of the students have been to Russia. \textbf{(individual comparison)} Whereas I, not as much \textbf{(event comparison)}. & Implausible \\ \hline
Outlier & The corrected sentence deviates from the illusory one by too many edits compared to the typical edit distance. & More musicians have listened to Apple Music than the amateur has. $\rightarrow$ Apple Music is more popular to musicians than it is to amateur music listeners. & Implausible (we label this category as ``implausible'' by convenience; interpretations in this category could be plausible by nature)  \\ \hline
Ungrammatical & Ungrammatical sentences that do not belong to the other categories. & More teenagers have used Tiktok than we have.$\rightarrow$ More teenagers than we \textbf{have, have} used TikTok. & Implausible \\ \hline
\end{longtable}

\subsection{Results}

\subsubsection*{Distribution of interpretations}

\begin{figure}[ht!]
    \centering
    \includegraphics[width=\linewidth]{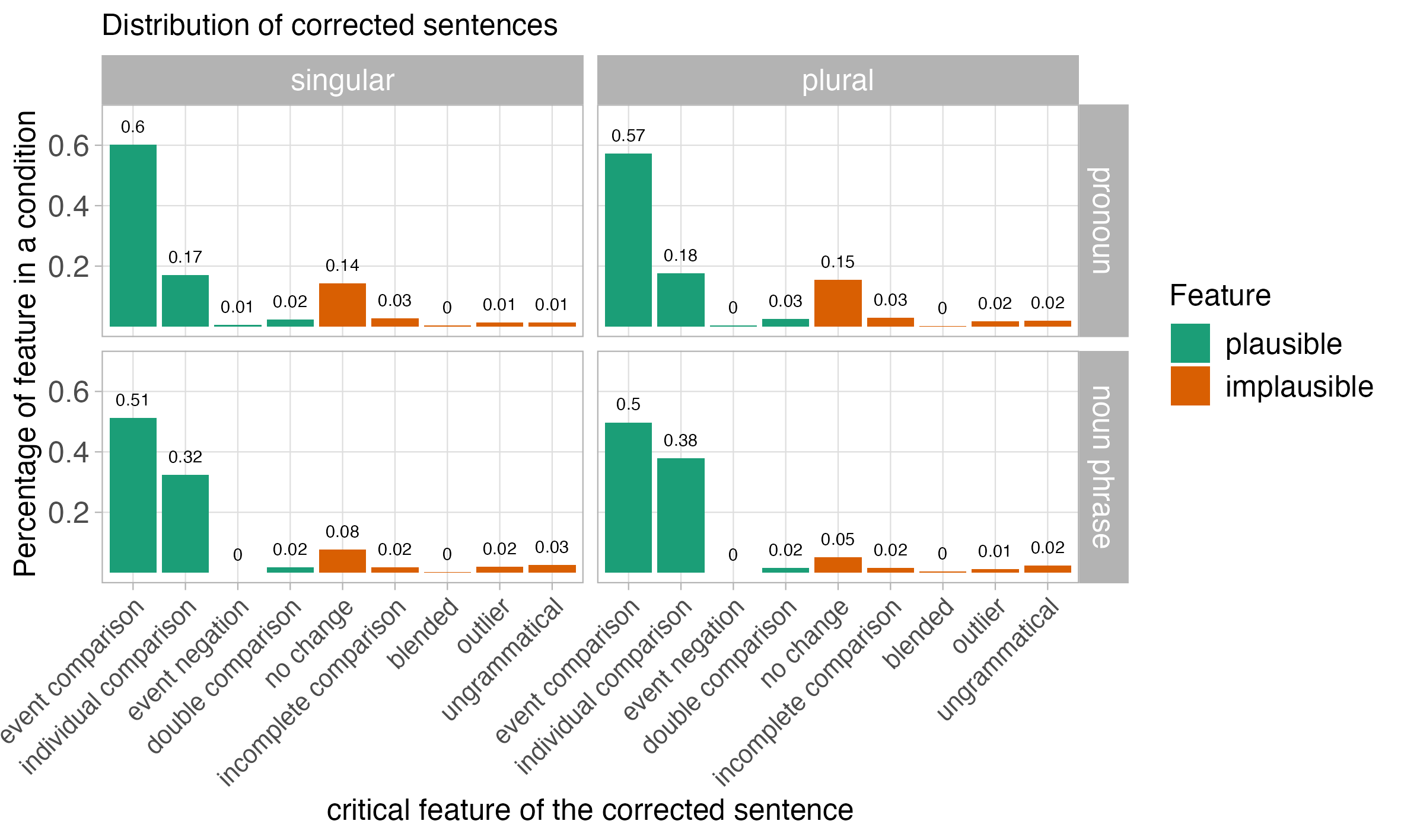}
    \caption{The percentage distribution of different interpretation categories of the corrected sentences in Experiment 2}
    \label{fig:exp2-qualitative}
\end{figure}

Figure \ref{fig:exp2-qualitative} represents the percentage of plausible versus implausible corrections across \textit{than}-clause subject form and number for 5,970 responses. Around 80\% of the corrections are plausible, suggesting that interpretations were drawn toward salient plausible (even if non-veridical) interpretations of the perceived CI sentence, in line with \textcite{paape2024linguistic} and \cite{zhang_comparative_2024}. 
It also represents the proportions of all interpretation categories of the corrected sentences. Across the four conditions, at least half of the corrections adjusted the usage of \textit{more} to convey a comparison between event frequencies. This ``event comparison'' feature is followed by an ``individual comparison'' feature where either the \textit{than} clause was fronted or the \textit{than}-clause subject structure was modified. The third most common category is \textit{no change}. The observation that a higher percentage of \textit{no change} appears in the pronoun condition is indicative of its higher acceptability in Experiment 1 -- comprehenders have a higher likelihood to regard CI sentences with a pronoun than-clause subject as acceptable and thus there is no need for change.

\subsubsection*{Word-level edit distance}

\begin{figure}[ht!]
    \centering
    \includegraphics[width=0.8\linewidth]{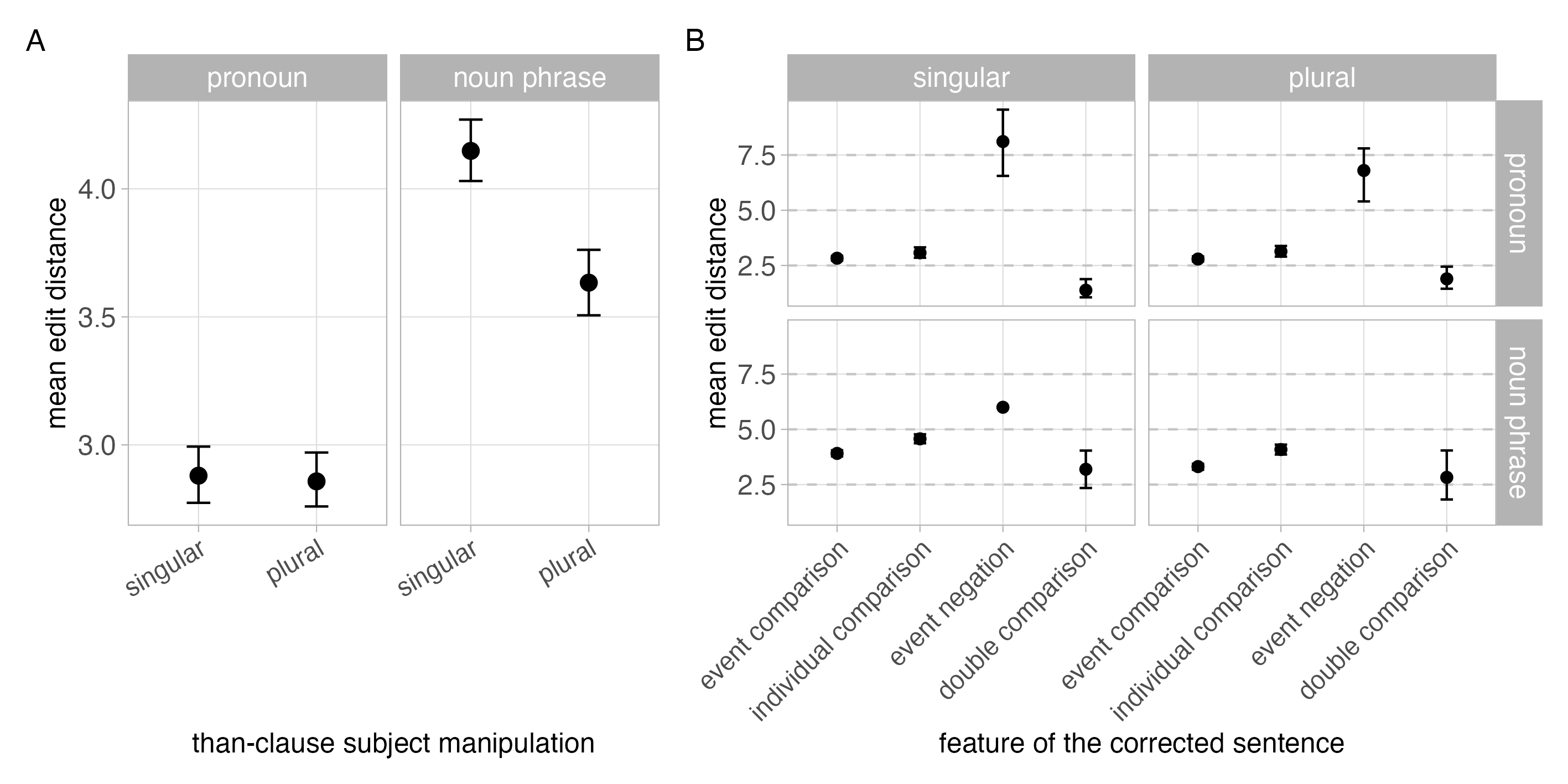}
    \caption{Mean DLD edit distance across the \textit{than}-clause subject manipulations (A) and plausible interpretations across the four conditions (B). (Error bars represent the 95\% bootstrapped confidence interval; the lack of data point for \textit{event negation} corrections in the plural noun phrase condition is due to the lack of actual relevant corrections.)}
    \label{fig:exp2-dist-by-feature}
\end{figure}

Focusing on plausible corrections (4,965 trials), Figure \ref{fig:exp2-dist-by-feature} reveals the average word-level Damerau-Levenshtein distance (DLD) between CI sentences and corrected ones, across \textit{than}-clause subject conditions (A) and interpretation categories (B). Focusing on Figure \ref{fig:exp2-dist-by-feature}A and connecting Experiment 2 to Experiment 1, the 2 x 2 conditions show a ranked correlation: a smaller mean edit distance for a condition is associated with a smaller difference in acceptability between illusory sentences in that condition and the corresponding acceptable control (Fig. \ref{fig:exp1-acceptability}).\footnote{Although edit distance in Fig.~\ref{fig:exp2-dist-by-feature} is a strong predictor of relative acceptability \textit{within} the pronoun and noun phrase conditions (our core finding), it also predicts a difference \textit{between} the pronoun and noun phrase conditions (noun phrase conditions have systematically higher edit distance than pronoun conditions) that we do not find in the acceptability judgments in Fig.~\ref{fig:exp1-acceptability}. We speculate that this difference may be driven by inflation of edit distances in the plural NP condition as a result of a common edit strategy in that condition: fronting the \textit{than} clause incurs a large edit distance relative to a (semantically equivalent) alternative deletion of \textit{the} (e.g., \textit{More lawyers have vacationed in Florida than judges have} vs. \textit{More lawyers than judges have vacationed in Florida.}). This fronting edit might be excessively penalized by the simple heuristics of DLD relative to its subjective probability in the minds of comprehenders, a possibility that we leave to future work.} This hypothesis is supported by a significant Pearson correlation between item-wise mean edit distance from plausible corrections and mean acceptability difference from acceptable controls ($r=.492\text{, }p<.001$).\footnote{The item-wise acceptability rating difference was calculated in three steps: (1) Take the standardized acceptability rating of all trials from Exp. 1 based on each participant; (2) Take the item-wise mean of acceptability across the six conditions exemplified in (\ref{ex:exp1-material}), which resulted in 30 instances of mean acceptability for the total six conditions; (3) Calculate the difference between the mean acceptability of each illusory condition and that of its corresponding control, resulting in 120 data points. Altogether there were 120 data points for both the DLD metric and the acceptability difference metric.} Focusing on Figure \ref{fig:exp2-dist-by-feature}B, ``event comparison'' corrections have a smaller mean DLD compared to ``individual comparison'', across all four conditions. ``Event negation'' has the largest DLD and ``double comparison'' the lowest.

\subsection{Discussion}
Experiment 2 revealed several recurring strategies to correct anomalous CI sentences, including adjusting the syntactic positions of \textit{more} and the \textit{than} clause. Our descriptive analyses reveal that the majority of corrections were toward plausible interpretations. \revise{The most common interpretations were ``event comparison'' and ``individual comparison''. We acknowledge that one limitation of the current design is that introducing the meaning of \textit{more} before the experiment might prime participants about possible editing strategies and limit them from more creative edits. But in general, the patterns in Figure \ref{fig:exp2-qualitative} align with previous findings with no such instructions and support the prevalence of the ``event comparison'' and ``individual comparison'' interpretations \parencite{christensen2016dead,zhang_comparative_2024}.} We operationalized the similarity between the anomalous and corrected sentences as their Damerau-Levenshtein word-level edit distance, thereby using DLD to approximate the noise magnitude under a random-edit noise model. The average edit distance in each illusion condition reflects its acceptability difference from the acceptable control in Experiment 1.
This result is predicted by noisy-channel theory: CI acceptability is negatively related to noise magnitude as measured by DLD, and thus should in turn be positively related to the likelihood term of Eq. \ref{eq:bayes}.
In the following section, we subject this prediction to a more fine-grained test by combining DLD with statistical language models to approximate all three terms of Eq. \ref{eq:bayes} (prior, likelihood, and evidence) in order to study the quantitative relationship between CI acceptability and the noisy-channel posterior.

\section{Statistical Modeling} \label{sec:intro_modeling_method}

\revise{In Section \ref{sec:stats_modeling_1}, we derive an approximation of the noisy channel posterior distribution by integrating probabilities from large language models with the sample of plausible corrections and their associated edit distances from Experiment 2. In Sections \ref{sec:stats_modeling_control} and \ref{sec:stats_modeling_regression}, we use regression models to test the capacity of this estimated posterior distribution to explain variance in CI illusion strength as measured by the acceptability data in Experiment 1, over and above independently motivated controls. To preview the results, we find that noisy-channel posterior probabilities of intended interpretations explains unique variance in CI illusion strength.}

\subsection{Linking the posterior distribution $p(s_i | s_p)$ to human acceptability judgments} \label{sec:stats_modeling_1}

Let $s_p$ be an anomalous CI sentence. We hypothesize that the acceptability of $s_p$ will be influenced by possible intended utterances $s_i \in \mathcal{S}$ under noisy-channel assumptions about the comprehension process.
\revise{In theory, the number of possible intended meanings behind $s_p$ in the set $\mathcal{S}$ could be infinite, making the exact posterior intractable to estimate. So we approximate the probabilities of the most frequent subset of \textit{plausible} corrections $\hat{\mathcal{S}}_{s_p}$ given by the human participants in Experiment 2 in response to $s_p$. In other words, we take participants' plausible corrections for a certain $s_p$ to be its $s_i$ candidates. These plausible corrections were those labeled in our analysis of Experiment 2 as ``event comparison'', ``individual comparison'', ``event negation'', and ``double comparison''.
Because a single judgment for $s_p$ is assumed to be influenced by potentially many $s_i$, a linking function aggregating over $p(s_i | s_p), s_i \in \hat{\mathcal{S}}_{s_p}$ is needed.}
Here we consider two such linking functions (max and mean):\footnote{We also tested a third option, the weighted mean $f_{\text{weighted}}$, which is a variant of $f_{\text{mean}}$ that modulates the influence on acceptability of likely alternatives in proportion to their posterior probability. Because the predictive performance of this variant was poor in exploratory analyses, we did not systematically evaluate it. See Appendix ~\ref{sec:appendix_exploratory_phase} for the results on an exploratory dataset. 
\begin{equation} \label{eq:posterior-weighted-mean}
    f_{\text{weighted}}(s_p) = \frac{1}{\sum_{s_i \in \hat{\mathcal{S}}_{s_p}} p(s_i|s_p)}\sum_{s_i \in \hat{\mathcal{S}}_{s_p}} p(s_i|s_p)^2
\end{equation}
}
\begin{equation} \label{eq:posterior-aggregate}
\begin{aligned}
    & f_{\text{max}}(s_p) = \max_{s_i \in \hat{\mathcal{S}}_{s_p}} p(s_i|s_p) \\
    & f_{\text{mean}}(s_p) =\frac{1}{|\hat{\mathcal{S}}_{s_p}|}\sum_{s_i \in \hat{\mathcal{S}}_{s_p}}p(s_i|s_p)
\end{aligned}
\end{equation}
The $f_{\text{max}}$ and $f_{\text{mean}}$ links represent distinct hypotheses about how the posterior influences acceptability, namely, via the single most likely interpretation (max) or via multiple likely interpretations (mean). A single-interpretation account could reflect a \textit{satisficing} mechanism consistent with prior reports in decision-making problems \parencite{simon1955behavioral,simon1972theories,simon1978rationality}: in searching for an optimal solution, people may stop at a good enough choice and not exhaust all options available. 
This view also aligns with evidence accumulation (or \textit{race}) models from the decision making literature \parencite{busemeyer1993decision,donkin2018response,tillman2020sequential}: multiple alternatives are activated in parallel and compete to be the first to reach some threshold, at which point the winner is selected as the decision.
Alternatively, comprehenders could consider multiple alternatives, track their probabilities simultaneously, and aggregate across these probabilities to render a judgment. This hypothetical pattern would be consistent with a parallel processing models of sentence comprehension \parencite{mcclelland1989sentence, mcrae2013constraint}. Each link instantiates a hypothesis about how a human will judge the acceptability of a given CI sentence $s_p$ on average.

According to the noisy-channel equation  (Eq.~\ref{eq:bayes}, i.e., Bayes rule applied to language comprehension), the posterior probability $p(s_i|s_p)$ can be calculated as the product of the prior probability $p(s_i)$ and the likelihood $p(s_p|s_i)$ divided by the evidence $p(s_p)$.
Therefore, to obtain values for the linking functions in Eq.~\ref{eq:posterior-aggregate} for a given $s_p$, we must compute (or estimate) these three terms on the right hand side of Eq. \ref{eq:bayes}.
We estimated the prior $p(s_i)$ and evidence $p(s_p)$ using sentence probabilities $p_m$ from a pretrained language model $m$, where $s_t$ is the $t^{\text{th}}$ word in sentence $s$:
\setlength{\jot}{20pt}
\begin{align}
    & p_m(s) = \prod_{t=1}^{T} p_m(s_{t} | s_{0...t-1}) \label{eq:P_lm}\\
    & \hat{p}(s_i) = p_m(s_i) \label{eq:prior} \\
    & \hat{p}(s_p) = p_m(s_p) \label{eq:evidence}
\end{align}
We estimated $\hat{p}(s_i)$ and $\hat{p}(s_p)$ using both GPT-2 Small (124 million parameters) \parencite{radford2019language} and OPT (1.3 billion parameters) \parencite{zhang2022opt}. We chose GPT-2 Small based on prior evidence of strong psychometric fit to human behavioral data during naturalistic language processing \parencite{oh2023does,shain2024large}. We chose an OPT model with a larger parameter count to verify that results do not depend critically on the choice of language model. \revise{We assume that these computational language models can distinguish grammatical and plausible intended messages $s_i$ and anomalous CI sentences $s_p$ by assigning a lower probability score to the ungrammatical sentences, with independent support showing that probabilities from models such as GPT-2 and -3 do not show CI illusion effects\parencite{zhang_can_2023}. Given no independent evidence for OPT in detecting the CI anomaly, we do not rule out the chance that OPT might track language statistics differently from GPT-2 Small. This is why comparing results from multiple models is crucial to deriving generalizable insights.} We used the \texttt{minicons} package \parencite{misra2022minicons} to compute the sentence probability from language models. 

The likelihood $p(s_p | s_i)$ is difficult to approximate using a language model because of its dependence on the intended utterance $s_i$ behind an observed utterance $s_p$, which is neither directly observable nor approximated by the model in any interpretable way.
Fortunately, \revise{Experiment 2 provides a distance metric between $s_i$ and $s_p$ that can approximate the likelihood empirically, following \textcite{meylan2023adults}. Specifically, the likelihood $\hat{p}(s_p|s_i)$ is assumed to be proportional to the exponentiated negative Damerau-Levenshtein edit distance (DLD), thereby enforcing an assumption that major corruptions occur less frequently in the channel than minor corruptions}:
\setlength{\jot}{20pt}
\begin{align}
    & \hat{p}(s_p|s_i) \propto e^{-\beta \cdot \text{DLD}(s_i,s_p)} \label{eq:likelihood}
\end{align}
The free parameter $\beta$ scales the likelihood: as $\beta$ approaches 0, all likelihoods approach 1. Here we do not explore $\beta$ and instead simply set it to 1.

This definition of $\hat{p}(s_p|s_i)$ has serious limitations.
It is not properly normalized (normalizing over the infinite set $\mathcal{S}$ is intractable). Nor does it discriminate the likelihood of different types of edits in ways that human minds likely do (e.g., based on acoustic, articulatory, morphosyntactic, or semantic similarity), and the exponentiality assumption is motivated by mathematical convenience rather than any cognitive consideration.
We use it because (1) it has the desirable property of monotonically decreasing the likelihood for more severe distortions, (2) its missing normalizing constant will be absorbed into regression models \revise{due to standardization of independent variables}, (3) it assigns a graded probability to any member of the open set of \{$s_i, s_p$\}, and (4) few alternatives exist because instantiating the likelihood term is an outstanding unsolved problem in computational linguistics, requiring in principle a complete working model of noisy-channel language comprehension before it can be computed exactly.

With these definitions in hand, we can define the approximate posterior directly as:

\setlength{\jot}{20pt}
\begin{align}
    & \hat{p}(s_i|s_p) = \frac{\hat{p}(s_i) \hat{p}(s_p|s_i)}{\hat{p}(s_p)} \label{eq:posterior-approx}
\end{align}

We substitute $\hat{p}(s_i|s_p)$ from Eq.~\ref{eq:posterior-approx} for $p(s_i|s_p)$ in Eq.~\ref{eq:posterior-aggregate} to yield two quantitative noisy-channel-based models for CI acceptability, one for each hypothesized link (max and mean). If CI effects are indeed a consequence of noisy-channel inference, there should be a positive relationship between one or more of these \revise{model-derived estimates of posterior probability} and the acceptability of CI sentences.
In other words, an increase in the posterior probability of one or more interpretations of a CI sentence should produce a proportional increase in its acceptability rating. \revise{There were 120 unique anomalous stimuli $s_p$ tested in Experiments 1 and 2 and each linking function assigns a value to each stimulus, totally 120 data points for each linking function.}

\subsection{Control variables}
\label{sec:stats_modeling_control}

Because natural language is a complex medium that can introduce hard-to-detect confounds \parencite{Smith2015-ip,Shain2023-qc}, we controlled statistically for variables known to influence sentence acceptability.
Our set of control variables was as follows.

First, we considered the Syntactic Log-Odds Ratio (SLOR) \parencite{lau2017grammaticality, pauls2012large} as a joint measure of the influence of contextual probability, word frequency, and sentence length.
SLOR is defined as follows, where $p_u(w)$ is the probability of word $w$ under unigram distribution $u$):
\begin{equation} \label{eq:slor}
    \text{SLOR}(s) = \frac{\log p_m(s)-\sum_{w \in s}\log p_u(w)}{|s|}
\end{equation}
In words, SLOR is the per-word difference between a sentence's probability (according to a statistical language model $m$) and its mean unigram probability.
SLOR has been argued to be a strong predictor of sentence acceptability \parencite{lau2017grammaticality, lau2020furiously}. For the language model, we used GPT-2 Small and OPT. Since the training data sets for GPT-2 Small and OPT are not publicly accessible, we estimated the unigram probability of a word based on word frequency in the English Wikipedia article dumps accessed on April 13, 2023, following the method in \textcite{lu2024can}. As was the case with our noisy-channel linking functions above, SLOR assigns a value to each of the 120 unique anomalous stimuli in Experiments 1 and 2.

In addition to SLOR, we controlled for the \textit{order} in which a stimulus occurred in a participant's reading list in Experiment 1 as a separate control variable. Since a participant read 94 sentences in Experiment 1, the order variable ranged from 1 to 94 and was standardized before entering the model.

Finally, we also controlled for the \textit{baseline} acceptability of the (non-anomalous) control sentence corresponding to a given anomalous sentence in the experiment, as a way of capturing item effects.

\subsection{Statistical model design} \label{sec:stats_modeling_regression}

We ran multiple mixed-effects Bayesian multilevel ordinal regression models \parencite{burkner2019ordinal} (via \textit{brms} \parencite{burkner2017brms} in R \parencite{r2024r}) to predict the CI acceptability ratings from Experiment 1 under each of our hypothesized linking functions $f \in \{f_{\text{max}}, f_{\text{mean}}\}$, over and above controls. The model design is shown in (\ref{eq:linear-regression}).

\begin{equation} \label{eq:linear-regression}
\begin{aligned}
    \text{Acceptability}(s_p) =\ & \beta_0 + \underbrace{\beta_1 \cdot f(s_p))}_{\text{noisy-channel}} +\\
    &\underbrace{\beta_2 \cdot \text{SLOR}(s_p) + \beta_3 \cdot \text{order}(s_p) + \beta_4 \cdot \text{baseline}(s_p)}_{\text{controls}} + (1\ |\ \text{participant}) + \epsilon
\end{aligned}
\end{equation}

The response variable $\text{Acceptability}(s_p)$ was the raw Likert scale acceptability score ranging from 1 to 7. 
We standardized all explanatory variables \parencite{gelman2007data} so that coefficients are in standard units.
All models had weakly informative priors: the intercepts and the population-level coefficients had a prior Gaussian distribution (i.e., $\text{Normal}(0,2)$) and the by-participant varying intercepts were given a weakly informative Half-Cauchy prior (i.e., $\text{Cauchy}(0,10)$), following the set up in Experiment 1. Each model had four sampling chains with 2000 iterations and the first 1000 iterations were taken as warmup. The sampling chains for all coefficients in all models had a $\hat{R}$ of 1, indicating successful convergence.

\subsection{Model comparison}

The goal of statistical modeling was to test whether noisy channel explains variation in CI acceptability over and above controls, and, if so, which noisy channel link was best supported.

We divided modeling into an exploration phase and a generalization phase. In the exploration phase, we tested the predictive performance of all three linking functions represented by \textsc{max}, \textsc{mean}, and \textsc{weighted mean} by predicting the acceptability of CI sentences in a published dataset in \textcite{zhang_comparative_2024} (857 trials). In the generalization phase, we investigated whether the statistical patterns held in the new dataset collected in Experiment 1 (9880 trials). We decided to drop the \textsc{weighted mean} from the generalization phase due to its poorer performance in the exploration phase compared to \textsc{mean}. For both phases, we examined whether GPT-2 Small and OPT generated similar patterns.

For each phase and each language model, we adopted a forward stepwise regression method by initiating multiple statistical models with the design in Table \ref{tab:structure-stat-model}. We started with a model with all three control predictors and incrementally added any of the critical posterior models until the statistical model no longer improved. \revise{We rely on the LOOIC score (leave-one-out information criterion) to determine models' predictive performance --- the smaller the LOOIC, the better the predicted values captured the distribution of the true values \parencite{vehtari2017practical}. The unique contribution of an explanatory variable is captured by the difference in LOOIC between a model that contains it as a fixed effect and another that does not, all other things being equal.}

\begin{table}[ht]
    \centering
    \begin{tabular}{|p{4cm}|p{12cm}|}
        \hline
         \textbf{Model name} & \textbf{Predictors}  \\
         \hline
         Base & = controls (SLOR, order, and baseline) + (1\ |\ participant)\\ \hline
         \textsc{max} & = controls + $f_{\text{max}}$ + (1\ |\ participant) \\ \hline
         \textsc{mean} & = controls + $f_{\text{mean}}$ + (1\ |\ participant) \\ \hline
         \textsc{max}, \textsc{mean}  & = controls + $f_{\text{max}}$ + $f_{\text{mean}}$ + (1\ |\ participant) \\ \hline
         \textsc{max}, \textsc{weighted}  & = controls + $f_{\text{max}}$ + $f_{\text{weighted}}$ + (1\ |\ participant) \\ \hline
         \textsc{mean}, \textsc{weighted}  & = controls + $f_{\text{mean}}$ + $f_{\text{weighted}}$ + (1\ |\ participant) \\ \hline
         \textsc{max}, \textsc{mean}, \textsc{weighted}  & = controls + $f_{\text{max}}$ + $f_{\text{mean}}$ + $f_{\text{weighted}}$ + (1\ |\ participant) \\ \hline
    \end{tabular}
    \caption{Statistical model names and definitions}
    \label{tab:structure-stat-model}
\end{table}

\subsection{Results} \label{sec:stats_model}

In the exploration phase, we fitted eight statistical models with the design in Table \ref{tab:structure-stat-model} and compared their predictive performance based on leave-one-out cross validation. The results show that the \textsc{mean} posterior variable outperformed \textsc{weighted mean} and \textsc{max}. Both \textsc{mean} and \textsc{weighted mean} positively predict CI acceptability, but the effect of \textsc{max} was minimal (see details in Appendix \ref{sec:appendix_exploratory_phase}). Since \textsc{weighted mean} can be viewed as a variant of \textsc{mean}, we chose the \textsc{mean} model as the main candidate to represent the hypothesis that CI acceptability is influenced by multiple plausible alternatives, in comparison with the \textsc{max} model which assumes that only the highest probability interpretation influences acceptability.
We turn now to results on the generalization dataset (Experiment 1 of this study).

\begin{figure}[htbp]
\centering
\includegraphics[width=0.95\linewidth]{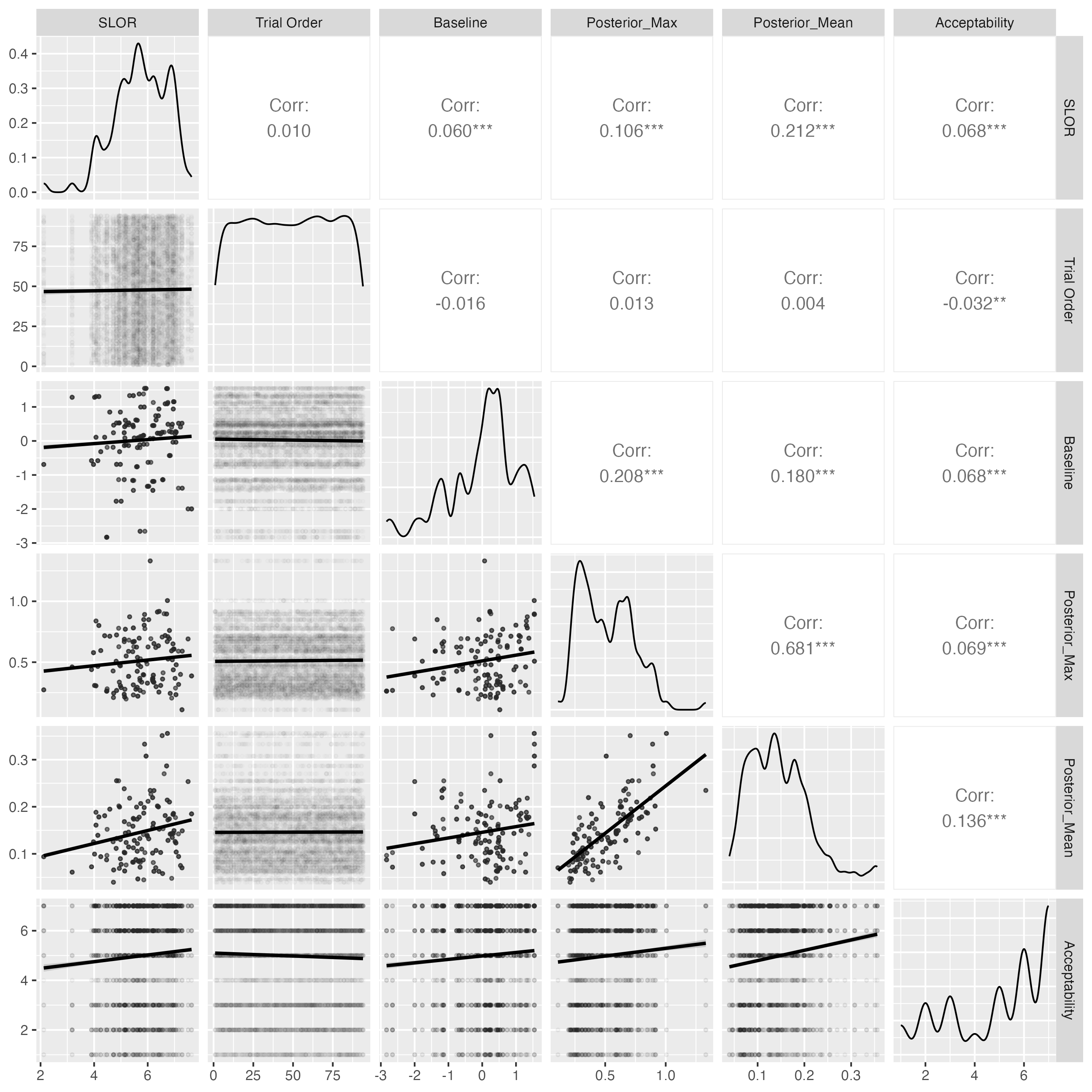}
\caption{\textbf{GPT-2 Small}: Pairwise correlation matrix for acceptability, SLOR, trial presentation order, acceptability of the baseline control sentence, and two posterior metrics.
The diagonal shows density plots for each metric; the cells in the upper triangle shows the Pearson correlation coefficient and the significance value; the cells in the lower triangle show scatter plots of two compared variables and a linear regression line with shadings representing standard error. All variables except for the baseline control rating are on their original scale.}
\label{fig:corr-measures-gpt2-hack-new-data}
\end{figure}

Starting with the results generated by GPT-2 Small, we first descriptively explored the correlations between the variables in our regression model.
Figure \ref{fig:corr-measures-gpt2-hack-new-data} shows that the CI acceptability significantly correlated with all explanatory variables positively, except for a significant negative correlation with the trial order. Multicollinearity existed between SLOR, the acceptability of the baseline control sentence, the \textsc{max} and \textsc{mean} posterior variables. We thus relied on statistical modeling to tease apart their contributions to CI acceptability.

Four mixed-effects Bayesian multilevel cumulative ordinal regression models were initiated to differentiate the effects of the \textsc{mean} and \textsc{max} posterior models by allowing neither, one, or both into the statistical models. We used the \texttt{loo} package \parencite{loo} in R to conduct Bayesian leave-one-out-cross-validation \parencite[LOO-CV][]{vehtari2017practical} to evaluate model performance, as shown in Table \ref{tab:looic_gpt2_hack_new_data}. \revise{A lower LOOIC indicates better fit. The best-fitting model was \textsc{max}, \textsc{mean}, followed by \textsc{mean}, followed by \textsc{max}, followed by \textsc{base}.}

\begin{table}[htbp]
\centering
\begin{adjustbox}{width=0.7\linewidth}
\begin{tabular}{lrrrr}
\hline
\textbf{model} & \textbf{LOOIC} & \textbf{SE} & \textbf{$\Delta$ LOOIC} & \textbf{$\Delta$ SE}  \\ \hline
\textsc{max}, \textsc{mean} & $29337.38$ & $183.98$ & $0.00$ & $0.00$ \\
\textsc{mean}      & $29358.70$ & $183.99$ & $21.32$ & $0.01$\\
\textsc{max}       & $29542.67$ & $184.37$ & $205.29$ & $0.39$\\
\textsc{base}      & $29585.08$ & $184.51$ & $247.70$ & $0.53$\\ \hline
\end{tabular}
\end{adjustbox}
\caption{\textbf{GPT-2 Small}: LOOIC (and its standard deviation) by model. (LOOIC = leave-one-out information criterion; smaller LOOIC indicates better fit.)} \label{tab:looic_gpt2_hack_new_data}
\end{table}

Table \ref{tab:coefficients_gpt2_hack_new_data} represents the estimated coefficients from all four statistical models. The three control variables show consistent effects: the \textit{SLOR} effect shows that the more probable a language model deems an anomalous sentence, the more acceptable the sentence is; the \textit{Trial order} effect shows that increased exposure to anomalous sentences decreases their acceptability; the \textit{Baseline} effect shows that CI sentences with more acceptable non-anomalous controls are rated more acceptable themselves. The \textsc{mean} posterior model positively predicts CI acceptability, consistent with the findings from the exploration phase and the predictions of the noisy-channel account of CI illusion effects. The \textsc{max} model was not strongly predictive in the exploration phase but shows a positive correlation with CI acceptability in the generalization phase. Still, \textsc{mean} numerically outperformed \textsc{max}, and when \textsc{mean} is present in the model, the coefficient for \textsc{max} becomes negative, suggesting that noisy-channel effects are best captured by the linking function ($f_{\text{mean}}$) that aggregates over multiple alternatives.

\begin{table}[htbp]
\centering
\begin{adjustbox}{width=\linewidth}
\begin{tabular}{lccccc}
\hline
\textbf{model} & \textbf{SLOR} & \textbf{Trial Order} & \textbf{Baseline} &
\textbf{$f_{\text{max}}$} & \textbf{$f_{\text{mean}}$} \\ \hline
\textsc{max}, \textsc{mean} & $0.097\,[0.059,\,0.137]$ & $-0.071\;[-0.110,\,-0.032]$ & $0.133\,[0.093,\,0.173]$ & $-0.125\;[-0.177,\,-0.072]$ & $0.390\,[0.335,\,0.446]$ \\
\textsc{mean}      & $0.103\,[0.064,\,0.142]$ & $-0.072\;[-0.112,\,-0.034]$ & $0.120\,[0.080,\,0.159]$ & ---                       & $0.305\,[0.264,\,0.346]$ \\
\textsc{max}       & $0.148\,[0.111,\,0.185]$ & $-0.073\;[-0.112,\,-0.035]$ & $0.142\,[0.102,\,0.182]$ & $0.127\,[0.087,\,0.166]$  & --- \\
\textsc{base}      & $0.158\,[0.119,\,0.195]$ & $-0.072\;[-0.110,\,-0.034]$ & $0.170\,[0.132,\,0.208]$ & ---                       & --- \\ \hline
\end{tabular}
\end{adjustbox}
\caption{\textbf{GPT-2 Small}: Estimated coefficients of explanatory variables.
The estimate is the mean of the posterior distributions of each explanatory variable, with the 95\% Bayesian credible interval in square brackets.} \label{tab:coefficients_gpt2_hack_new_data}
\end{table}

Results based on OPT converged with what is reported here and support the superiority of the linking function $f_{\text{mean}}$ over $f_{\text{max}}$ in predicting the variance in CI illusion strength. Details can be found in Appendix ~\ref{sec:appendix_opt_generalization}.

\subsection{Discussion}

We integrated human behavioral data with statistical language models and Bayesian regression models to test how well the noisy-channel posterior predicts human acceptability ratings for CI sentences. We found that a linking function which averaged the posterior probabilities of multiple likely interpretations ($f_{\text{mean}}$ in Eq. \ref{eq:posterior-aggregate}) explained variance in CI acceptability over strong controls and outperformed an alternative linking function that only considered the most likely interpretation ($f_{\text{max}}$).
This superiority of $f_{\text{mean}}$ replicated in two independent datasets and under two different language models. This result supports the hypothesis that rational Bayesian inference in comprehension explains the comparative illusion and indicates that comprehenders' acceptability judgments are influenced by multiple plausible interpretations of CI sentences, not just the most probable one under the posterior.

\section{General Discussion} \label{sec:discussion}

Language illusions are a critical window into the mental processes that support human language comprehension.
In this study, we have joined recent work \parencite{paape2024linguistic,zhang_comparative_2024} in arguing that one well-studied linguistic illusion---the \textit{comparative illusion}, e.g., \textit{More people have been to Russia than I have} \parencite{montalbetti1984after}---\revise{is explained by} a general Bayesian theory of rational language comprehension: noisy-channel inference \parencite{gibson_rational_2013, levy_noisy-channel_2008}.
Our study links a key behavioral indicator of illusion strength---acceptability judgments---to the core mathematical construct of noisy-channel theory, the posterior probability distribution.
We have thus gone substantially beyond prior work on this question \parencite{paape2024linguistic,zhang_comparative_2024}, which has to-date focused on gathering behavioral data in support of noisy-channel predictions but has not studied the quantitative fit of the mathematical constructs of the noisy-channel model to behavioral data.
In particular, we have argued that comprehenders of anomalous CI sentences consider a range of plausible interpretations in proportion to their posterior probability under channel noise, and render an acceptability judgment that is well approximated by the average probability of multiple interpretational candidates. Our empirical results thus show that the graded illusion strength in CI sentences is indeed explained by a quantitative model of the posterior probabilities of plausible alternatives, which we approximated by integrating open-ended sentence probabilities derived from computational language models \parencite{radford2019language,zhang2022opt} with open-ended human behavioral data (a ``sentence correction'' task in Experiment 2). 

In our study, Experiment 1 replicated the graded illusion strength reported in prior smaller scale studies \parencite{zhang_comparative_2024} under a nearly order-of-magnitude increase in sample size.
We confirmed that different types of CIs gave rise to distinctive profiles of acceptability caused by simple changes in the \textit{than}-clause subjects (\ref{ex:ci-four-exs}) \parencite{oconnor2013,oconnor2015comparative,wellwood2018anatomy}. Experiment 2 asked human participants to give the fewest edits needed to a CI sentence to make it have a plausible meaning, based on which we obtained an empirical sample of plausible interpretations and treated them as intended messages in the noisy-channel model (Eq.~\ref{eq:bayes}). The emerging interpretation categories (e.g., ``event comparison'' and ``individual comparison'') align with previous literature \parencite{christensen2016dead,kelley2018more,oconnor2013,oconnor2015comparative,wellwood2018anatomy}. We also obtained the Damerau-Levenshtein (DL) word-level edit distance from the perceived CI stimulus to the plausible interpretations. Using Bayes' rule (Eq.~\ref{eq:bayes}), we estimated the noisy channel posterior probability of intended messages for CI stimuli by integrating sentence probabilities of $s_i$ and $s_p$ from statistical language models (GPT-2 and OPT) with the approximate noise likelihood term derived from the DL distance obtained in Experiment 2 (Eqs.~\ref{eq:prior}--\ref{eq:likelihood}). We then approximated candidate linking functions (Eq.~\ref{eq:posterior-aggregate}) between the estimated noisy-channel posterior probability $\hat{p}(s_i|s_p)$ and the CI acceptability data, using Bayesian regression to test the contribution of these noisy-channel linking functions to explaining the acceptability data from Experiment 1 over and above strong controls.
We considered two possible links, one (\textsc{mean}) that averaged over the posterior probabilities of all plausible corrections of a CI sentence given by participants in Experiment 2, and one (\textsc{max}) that only considered the posterior probability of the most likely interpretation under the noisy-channel posterior. Results showed that the \textsc{mean} link positively predicted CI acceptability over and above controls and outperformed the \textsc{max} link. In other words, highly probable plausible alternatives of the CI stimuli drive up their illusion strength as indicated by increased acceptability ratings. This pattern replicated across the two computational language models (GPT-2 Small and OPT) and two disjoint CI acceptability datasets (the public dataset in \textcite{zhang_comparative_2024} and the dataset in Experiment 1).

\revise{Our study shows that variation in illusion strength across diverse CI sentences---largely caused by morphosyntactic differences of the \textit{than}-clause subjects---can be partially explained by noisy-channel inference. Our theory offers a link between surface linguistic structure in the illusion stimuli and human comprehension behavior.
Although we have stated our theory in general probabilistic terms, this probabilistic framework is compatible with possible fine-grained influences of context not explicitly considered here.
For example, differences in register (formal vs.\ informal) or modality (spoken vs.\ written) plausibly shape the strength of illusion effects.
For example, one possibility raised by a reviewer is that pronominal subjects may be more susceptible to CI because they more closely resemble conversational usage.
Although we cannot verify this impression empirically based on the data at hand, if it were true, one way our theory could account for it would be to predict that the probability of corruption (and thus, the probability of making a non-literal meaning inference) is higher in spontaneous conversation, a conjecture that is in line with known patterns of speech errors \parencite{chafe_integration_1982,dahan2019language,garrett_analysis_1975,levelt1983monitoring}.} 
Overall, this work joins a growing literature showing that various types of language illusions can be explained by the noisy-channel theory \parencite{paape2024linguistic,qian2023comprehenders,ryskin_agreement_2021,zhang_noisy-channel_2023,zhang2024rational}. This interpretation aligns with \textit{rational} theories of cognition \parencite{anderson1990adaptive,chater1999ten}, according to which perceptual systems (potentially including language comprehension) approximate Bayes-optimal integration of prior knowledge with (potentially noisy) sensory evidence. This view recasts language illusions from revealing failures of the comprehension system to revealing its proper functioning under the influence of prior knowledge, in line with similar arguments about illusions in other domains of cognition \parencite{von-Helmholtz1924-el,Gigerenzer1991-pb,Tenenbaum2001-iy,Griffiths2006-cv,Gershman2012-ax,Born2020-ic}. 

\revise{Our CI findings may also bear on ``good-enough'' theories of language processing \parencite{christianson_when_2016,christianson2006younger,ferreira2002good,ferreira2007good, karimi_good-enough_2016, paape2024linguistic}, which attempt to explain patterns of human language use by positing fast (and potentially errorful) strategies that might help comprehenders minimize processing effort while still being accurate enough for their current communicative needs.
Good-enough processing is thus a family of \textit{algorithmic}-level \parencite{marr2010vision} hypotheses about the kinds and contents of mental representations and processes that underlie language comprehension, and how biases introduced by these processes can lead to illusion phenomena and other systematic comprehension errors.
As stressed in section~\ref{sect:background}, noisy-channel theory as understood here and elsewhere \parencite{levy_noisy-channel_2008,gibson_rational_2013} is by contrast a \textit{computational}-level theory of Bayes-optimal sentence comprehension, which makes few algorithmic-level commitments.
Because these two classes of theories target different levels of analysis, they are potentially compatible.
For example, heuristic strategies from the good-enough processing literature may be part of the mechanism by which the comprehender (perhaps inadvertently) approximates the behavior that would be expected if they were performing Bayes-optimal inference over a noisy channel.}

\revise{Critically, however, the hypothesized cause of CI effects differs in substance between the two views.
Under good-enough processing, CI derives ultimately from minimizing comprehension effort under a particular set of assumptions about the effort required by different mental processes.
Under noisy-channel inference, however, CI effects are \textit{rational} with specific quantitative predictions from the Bayesian inference framework: they reflect the larger posterior probability assigned to CI sentences by a Bayesian comprehender, without regard to any constraints on the comprehender's cognitive resources.
Our view thus implies that CI effects can be explained by the mathematical structure of problem that the comprehension system is solving (determining what the speaker meant), irrespective of the processing mechanisms that give rise to this inference.
Unique support for this view comes from our demonstration that graded variation in posterior probability tracks graded variation in the acceptability of CI sentences.
The full pattern of this graded variation does not currently have an algorithmic-level (e.g., good-enough) explanation.}   

In addition to the key theoretical finding above, our results have methodological implications, converging with prior evidence \parencite{lau2017grammaticality,lau2020furiously} that (normalized) sentence probability under a statistical language model is a good predictor of acceptability.
This suggests that acceptability judgments are relevant to a richer space of questions than the (binary) grammaticality and felicity distinctions that originally motivated their development as a method \parencite{Gibson2010-ft}. Nevertheless, we found no adaptation or satiation effects where anomalous sentences could become more acceptable as the experiment unfolds \parencite[cf.][]{lu2024linguistic, snyder2000experimental}. On the contrary, the statistical models suggest that CI sentences become less acceptable as the experiment unfolds. Given that comprehenders of CI sentences have been reported to diagnose the anomaly \parencite{paape2024linguistic}, we speculate that the acceptability score goes down because participants become better at detecting the anomaly. \revise{This effect should be replicated with future studies and potentially be explained in a noisy-channel framework by allowing the posterior to update incrementally throughout the task.}

In general, the statistical regression results show that the (normalized) probability of the target sentence, its presentation order in the experiment and the posterior probabilities of plausible alternatives independently explain variance in CI acceptability. This finding supports a reinterpretation of the acceptability task as capturing real-time probabilistic sentence processing, similar to other behavioral and neural measures such as reading times \parencite{levy2008expectation}, eye movement \parencite{levy2009eye}, and event-related potentials \parencite{ryskin2021erp}. This insight may unify diverse prior evidence that sentence acceptability is determined not just by grammaticality, truthfulness, and felicity \parencite{schutze1996empirical, tonhauser2015empirical} but also domain-general cognitive factors such as memory constraints and executive functions \parencite[e.g.,][]{ferreira1991recovery, gibson1999memory, hofmeister2010cognitive, hofmeister2012islands}, all of which may modulate the probability of an observed sentence and its possible interpretations.


A remaining puzzle is that acceptability was better predicted by the \textsc{mean} linking function compared to \textsc{max}.
\revise{We discuss several implications here. \textit{First}, although we have tentatively interpreted the superiority of the \textsc{mean} linking function to suggest that multiple alternatives are considered simultaneously  \parencite[consistent with parallel processing views of sentence processing][]{altmann1998ambiguity,mcclelland1989sentence,mcrae2013constraint,Rasmussen2018-tq,hahn2022resource}, we cannot rule out the possibility that comprehenders instead sample single interpretations probabilistically, since this would have a similar aggregate effect on our measurements. We leave this question to future research. \textit{Second}, our theory leaves unspecified the fine-grained temporal dynamics by which alternative interpretations are considered, weighed, and accepted or rejected. These dynamics may not be amenable to theorizing at the computational level, and likely require algorithmic-level commitments about the underlying process. \textit{Third}, the difference between \textsc{mean} and \textsc{max} may arise from limitations of our approximation of the posterior probability: a \textit{post hoc} analysis shows that for many CI stimuli in Experiment 2, the alternatives with the highest posterior probability calculated through Eq.~\ref{eq:posterior-approx} had the ``double comparison'' meaning, even though the empirical frequency of that meaning category is low (Fig.~\ref{fig:exp2-qualitative}). This could be because the noise model (Eq.~\ref{eq:likelihood}) assigns a high likelihood for the ``double comparison'' interpretation (i.e., a deletion of \textit{more} in the adverbial position), making it the highest noise likelihood among all edits. This high noise likelihood in turn drove up the prior probabilities assigned by the statistical language models. If this explanation is on the right track, we would expect that a more sophisticated noise model could improve the fit of the \textsc{max} link. Future studies should delineate more mathematically precise and cognitively legitimate linking functions between posterior probabilities of alternatives and behavioral/neural measurements of noisy-channel processing.}

\revise{One further limitation lies in how we operationalized the noise model $p(s_p|s_i)$ in Bayes' rule (Eq.~\ref{eq:bayes}), which captures comprehenders' implicit knowledge of how noise could corrupt an intended message. This quantity is difficult to estimate for technical reasons, and our attempt to do so using explicit human edits in Experiment 2 has many limitations, among which was the assumption that human edits convey information about their (implicit and subjective) noise distribution. The reason why we treat corrected sentences from Experiment 2 as plausible CI interpretations is that we assume the actual mental alternatives for CI sentences are several words away in the linguistic form. This assumption itself is of theoretical interest because real comprehension might involve (i) mental reorganization of words that go beyond local alternatives, and/or (ii) creative interpretations that do not have the same lexicons as the perceived sentence. If true, the current implementation in Experiment 2 would underestimate the linguistic diversity of plausible alternatives. Future research should work on a more precise understanding of the real inferential strategies people use to rescue the meaning of a noise corrupted sentence and go beyond simple word-level edits as noise models in the noisy-channel framework.}

\revise{Finally, the noisy-channel framework should in principle explain why the illusion is stronger in the elided comparative structure (\ref{ex:ci-sentence}) compared to the fully fleshed out structure (\ref{ex:ci-sentence-expand}). We hypothesize that the comparative information in the elided sentence is unspecific and invites rational inference to reconstruct plausible comparison; the expanded sentence conveys a clear mismatch of comparative information and this clear anomalous signal inhibits further inference. We leave this comparison to future work.}


\section{Conclusion} \label{sec:conclusion}
We offer evidence that converges with recent work in supporting the hypothesis that comparative illusions can be explained by information-theoretic noisy-channel theories of language comprehension. By integrating statistical language models with human behavioral data, we construct a mathematical model to represent the posterior probabilities of plausible alternatives and show that this model can account for previously unexplained gradience in comparative illusion strength. This work joins a growing body of literature to supporting noisy-channel inference as a model of human language comprehension \parencite[][a.o.]{brehm2021probabilistic,chen2023effect,gibson2017don,poliak2025,ryskin2018comprehenders, ryskin2021erp,zhang_memory-based_2024}.

\section*{Declaration of competing interests}

The authors have no competing interests.

\printbibliography

@ARTICLE{Weiss2002-lx,
  title        = {Motion illusions as optimal percepts},
  author       = {Weiss, Yair and Simoncelli, Eero P and Adelson, Edward H},
  journaltitle = {Nat. Neurosci.},
  publisher    = {Springer Science and Business Media LLC},
  volume       = {5},
  issue        = {6},
  pages        = {598--604},
  date         = {2002-06-20},
  urldate      = {2025-06-28},
  language     = {en}
}

@ARTICLE{de-Vignemont2005-uw,
  title        = {Bodily illusions modulate tactile perception},
  author       = {de Vignemont, Frédérique and Ehrsson, Henrik H and Haggard,
                  Patrick},
  journaltitle = {Curr. Biol.},
  publisher    = {Elsevier BV},
  volume       = {15},
  issue        = {14},
  pages        = {1286--1290},
  date         = {2005-07-26},
  urldate      = {2025-06-28},
  language     = {en}
}

@BOOK{von-Helmholtz1924-el,
  title     = {Treatise on physiological optics},
  author    = {von Helmholtz, H},
  publisher = {Optical Society of America},
  location  = {Rochester},
  date      = {1924}
}

@INBOOK{Gigerenzer1991-pb,
  title     = {On Cognitive Illusions and Rationality},
  author    = {Gigerenzer, Gerd},
  booktitle = {Probability and Rationality},
  publisher = {BRILL},
  pages     = {225--249},
  date      = {1991-01-01},
  urldate   = {2025-10-29},
  language  = {en}
}

@ARTICLE{Smith2015-ip,
  title        = {Regression-based estimation of {ERP} waveforms: {I}. The
                  {rERP} framework: {rERPs} {I}},
  author       = {Smith, Nathaniel J and Kutas, Marta},
  journaltitle = {Psychophysiology},
  publisher    = {Wiley},
  volume       = {52},
  issue        = {2},
  pages        = {157--168},
  date         = {2015-02},
  language     = {en}
}

@ARTICLE{Geisler2002-xd,
  title        = {Illusions, perception and Bayes},
  author       = {Geisler, Wilson S and Kersten, Daniel},
  journaltitle = {Nat. Neurosci.},
  publisher    = {Springer Science and Business Media LLC},
  volume       = {5},
  issue        = {6},
  pages        = {508--510},
  date         = {2002-06},
  language     = {en}
}

@ARTICLE{Yang2021-mv,
  title        = {Human visual motion perception shows hallmarks of Bayesian
                  structural inference},
  author       = {Yang, Sichao and Bill, Johannes and Drugowitsch, Jan and
                  Gershman, Samuel J},
  journaltitle = {Sci. Rep.},
  publisher    = {Springer Science and Business Media LLC},
  volume       = {11},
  issue        = {1},
  pages        = {3714},
  date         = {2021-02-12},
  urldate      = {2025-06-28},
  language     = {en}
}

@ARTICLE{Born2020-ic,
  title        = {Illusions, delusions, and your backwards Bayesian brain: A
                  biased visual perspective},
  author       = {Born, Richard T and Bencomo, Gianluca M},
  journaltitle = {Brain Behav. Evol.},
  publisher    = {S. Karger AG},
  volume       = {95},
  issue        = {5},
  pages        = {272--285},
  date         = {2020},
  urldate      = {2025-06-28},
  language     = {en}
}

@ARTICLE{Gibson2010-ft,
  title        = {Weak quantitative standards in linguistics research},
  author       = {Gibson, Edward and Fedorenko, Evelina},
  journaltitle = {Trends Cogn. Sci.},
  volume       = {14},
  issue        = {6},
  pages        = {233--234},
  date         = {2010-06},
  urldate      = {2024-03-14},
  language     = {en}
}

@ARTICLE{Gershman2019-wa,
  title        = {What does the free energy principle tell us about the brain?},
  author       = {Gershman, Samuel J},
  journaltitle = {Neurons, Behavior, Data analysis, and Theory},
  publisher    = {The Neurons Behavior Data Analysis and Theory collective},
  volume       = {2},
  issue        = {3},
  pages        = {1--10},
  date         = {2019-10-25},
  urldate      = {2025-06-11},
  language     = {en}
}

@ARTICLE{Rasmussen2018-tq,
  title        = {Left-Corner Parsing With Distributed Associative Memory
                  Produces Surprisal and Locality Effects},
  author       = {Rasmussen, Nathan E and Schuler, William},
  journaltitle = {Cogn. Sci.},
  volume       = {42},
  issue        = {S4},
  pages        = {1009--1042},
  date         = {2018},
  urldate      = {2022-11-10},
  language     = {en}
}

@ARTICLE{Shain2023-qc,
  title        = {No evidence of theory of mind reasoning in the human language
                  network},
  author       = {Shain, Cory and Paunov, Alexander and Chen, Xuanyi and Lipkin,
                  Benjamin and Fedorenko, Evelina},
  journaltitle = {Cereb. Cortex},
  volume       = {33},
  issue        = {10},
  pages        = {6299--6319},
  date         = {2023-05-09},
  urldate      = {2023-05-17},
  language     = {en}
}

@ARTICLE{Goh2023-uq,
  title        = {The perception of silence},
  author       = {Goh, Rui Zhe and Phillips, Ian B and Firestone, Chaz},
  journaltitle = {Proc. Natl. Acad. Sci. U. S. A.},
  volume       = {120},
  issue        = {29},
  pages        = {e2301463120},
  date         = {2023-07-18},
  urldate      = {2023-07-10}
}

@ARTICLE{Pouget2013-se,
  title        = {Probabilistic brains: knowns and unknowns},
  shorttitle   = {Probabilistic brains},
  author       = {Pouget, Alexandre and Beck, Jeffrey M and Ma, Wei Ji and
                  Latham, Peter E},
  journaltitle = {Nat. Neurosci.},
  volume       = {16},
  issue        = {9},
  pages        = {1170--1178},
  date         = {2013-09},
  urldate      = {2023-09-05}
}

@ARTICLE{Nour2015-wt,
  title        = {Perception, illusions and Bayesian inference},
  author       = {Nour, Matthew M and Nour, Joseph M},
  journaltitle = {Psychopathology},
  publisher    = {S. Karger AG},
  volume       = {48},
  issue        = {4},
  pages        = {217--221},
  date         = {2015-08-08},
  language     = {en}
}

@ARTICLE{Tenenbaum2011-ev,
  title        = {How to grow a mind: statistics, structure, and abstraction},
  author       = {Tenenbaum, Joshua B and Kemp, Charles and Griffiths, Thomas L
                  and Goodman, Noah D},
  journaltitle = {Science},
  publisher    = {American Association for the Advancement of Science (AAAS)},
  volume       = {331},
  issue        = {6022},
  pages        = {1279--1285},
  date         = {2011-03-11},
  language     = {en}
}

@ARTICLE{Herz2001-vw,
  title        = {The influence of verbal labeling on the perception of odors:
                  evidence for olfactory illusions?},
  author       = {Herz, R S and von Clef, J},
  journaltitle = {Perception},
  publisher    = {SAGE Publications},
  volume       = {30},
  issue        = {3},
  pages        = {381--391},
  date         = {2001},
  urldate      = {2025-06-28},
  language     = {en}
}

@ARTICLE{Gershman2012-ax,
  title        = {Multistability and Perceptual Inference},
  author       = {Gershman, Samuel J and Vul, Edward and Tenenbaum, Joshua B},
  journaltitle = {Neural Computation},
  volume       = {24},
  pages        = {1--24},
  date         = {2012}
}

@INBOOK{Griffiths2001-rr,
  title     = {Bayesian models of cognition},
  author    = {Griffiths, Thomas L and Kemp, Charles and Tenenbaum, Joshua B},
  editor    = {Sun, Ron},
  booktitle = {The Cambridge Handbook of Computational Psychology},
  publisher = {Cambridge University Press},
  location  = {Cambridge},
  volume    = {753},
  pages     = {59--100},
  date      = {2001-01-01}
}

@INPROCEEDINGS{Tenenbaum2001-iy,
  title     = {The rational basis of representativeness},
  author    = {Tenenbaum, Joshua B and Griffiths, Thomas L},
  booktitle = {Proceedings of the 23rd annual conference of the Cognitive
               Science Society},
  date      = {2001}
}

@ARTICLE{Griffiths2006-cv,
  title        = {Optimal predictions in everyday cognition},
  author       = {Griffiths, Thomas L and Tenenbaum, Joshua B},
  journaltitle = {Psychol. Sci.},
  publisher    = {SAGE Publications},
  volume       = {17},
  issue        = {9},
  pages        = {767--773},
  date         = {2006-09},
  urldate      = {2025-10-29},
  language     = {en}
}

@ARTICLE{Gershman2015-jf,
  title        = {Computational rationality: A converging paradigm for
                  intelligence in brains, minds, and machines},
  author       = {Gershman, Samuel J and Horvitz, Eric J and Tenenbaum, Joshua B},
  journaltitle = {Science},
  publisher    = {American Association for the Advancement of Science (AAAS)},
  volume       = {349},
  issue        = {6245},
  pages        = {273--278},
  date         = {2015-07-17},
  language     = {en}
}

@book{zhang2024rational,
  title={The Rational Processing of Language Illusions},
  author={Zhang, Yuhan},
  year={2024},
  publisher={Harvard University}
}

@article{qian2023comprehenders,
  title={Comprehenders’ error correction mechanisms are finely calibrated to language production statistics},
  author={Qian, Peng and Levy, Roger Philip},
  year={2023},
  publisher={PsyArXiv}
}

@book{anderson1990adaptive,
  title={The Adaptive Character of Thought},
  author={Anderson, John Robert},
  year={1990},
  publisher={Psychology Press}
}

@article{chater1999ten,
  title={Ten years of the rational analysis of cognition},
  author={Chater, Nick and Oaksford, Mike and Chater, Nick and Oaksford, Mike},
  journal={Trends in cognitive sciences},
  volume={3},
  number={2},
  pages={57--65},
  year={1999},
  publisher={Elsevier}
}

@article{levy2009eye,
  title={Eye movement evidence that readers maintain and act on uncertainty about past linguistic input},
  author={Levy, Roger and Bicknell, Klinton and Slattery, Tim and Rayner, Keith},
  journal={Proceedings of the national academy of sciences},
  volume={106},
  number={50},
  pages={21086--21090},
  year={2009},
  publisher={National Academy of Sciences}
}

@book{parker2025illusion,
  title={Linguistic illusions: A case study on agreement attraction},
  author={Parker, Daniel},
  year={2025},
  publisher={Cambridge University Press}
}

@incollection{mcrae2013constraint,
  title={Constraint-based models of sentence processing},
  author={McRae, Ken and Matsuki, Kazunaga},
  booktitle={Sentence processing},
  pages={51--77},
  year={2013},
  publisher={Psychology Press}
}

@article{mcclelland1989sentence,
  title={Sentence comprehension: A parallel distributed processing approach},
  author={McClelland, Jay L and St. John, Mark and Taraban, Roman},
  journal={Language and cognitive processes},
  volume={4},
  number={3-4},
  pages={SI287--SI335},
  year={1989},
  publisher={Taylor \& Francis}
}

@article{busemeyer1993decision,
  title={Decision field theory: a dynamic-cognitive approach to decision making in an uncertain environment.},
  author={Busemeyer, Jerome R and Townsend, James T},
  journal={Psychological review},
  volume={100},
  number={3},
  pages={432},
  year={1993},
  publisher={American Psychological Association}
}

@article{sanford2002depth,
  title={Depth of processing in language comprehension: Not noticing the evidence},
  author={Sanford, Anthony J and Sturt, Patrick},
  journal={Trends in cognitive sciences},
  volume={6},
  number={9},
  pages={382--386},
  year={2002},
  publisher={Elsevier}
}

@article{ferreira2002good,
  title={Good-enough representations in language comprehension},
  author={Ferreira, Fernanda and Bailey, Karl GD and Ferraro, Vittoria},
  journal={Current directions in psychological science},
  volume={11},
  number={1},
  pages={11--15},
  year={2002},
  publisher={SAGE Publications Sage CA: Los Angeles, CA}
}

@article{ferreira2003misinterpretation,
  title={The misinterpretation of noncanonical sentences},
  author={Ferreira, Fernanda},
  journal={Cognitive psychology},
  volume={47},
  number={2},
  pages={164--203},
  year={2003},
  publisher={Elsevier}
}

@article{ferreira2007good,
  title={The ‘good enough’approach to language comprehension},
  author={Ferreira, Fernanda and Patson, Nikole D},
  journal={Language and linguistics compass},
  volume={1},
  number={1-2},
  pages={71--83},
  year={2007},
  publisher={Wiley Online Library}
}

@article{zhang2022opt,
  title={Opt: Open pre-trained transformer language models},
  author={Zhang, Susan and Roller, Stephen and Goyal, Naman and Artetxe, Mikel and Chen, Moya and Chen, Shuohui and Dewan, Christopher and Diab, Mona and Li, Xian and Lin, Xi Victoria and others},
  journal={arXiv preprint arXiv:2205.01068},
  year={2022}
}

@book{gelman2007data,
  title={Data analysis using regression and multilevel/hierarchical models},
  author={Gelman, Andrew and Hill, Jennifer},
  year={2007},
  publisher={Cambridge university press}
}

@book{marr2010vision,
  title={Vision: A computational investigation into the human representation and processing of visual information},
  author={Marr, David},
  year={1982/2010},
  publisher={MIT press}
}

@article{donkin2018response,
  title={Response times and decision-making},
  author={Donkin, Christopher and Brown, Scott D},
  journal={Stevens’ handbook of experimental psychology and cognitive neuroscience},
  volume={5},
  pages={349--377},
  year={2018},
  publisher={John Wiley \& Sons, Inc.}
}

@article{simon1978rationality,
  title={Rationality as process and as product of thought},
  author={Simon, Herbert A},
  journal={The American economic review},
  volume={68},
  number={2},
  pages={1--16},
  year={1978},
  publisher={JSTOR}
}

@article{simon1955behavioral,
  title={A behavioral model of rational choice},
  author={Simon, Herbert A},
  journal={The quarterly journal of economics},
  pages={99--118},
  year={1955},
  publisher={JSTOR}
}

@article{brehm2021probabilistic,
  title={Probabilistic online processing of sentence anomalies},
  author={Brehm, Laurel and Jackson, Carrie N and Miller, Karen L},
  journal={Language, Cognition and Neuroscience},
  volume={36},
  number={8},
  pages={959--983},
  year={2021},
  publisher={Taylor \& Francis}
}

@article{gibson2017don,
  title={Don’t underestimate the benefits of being misunderstood},
  author={Gibson, Edward and Tan, Caitlin and Futrell, Richard and Mahowald, Kyle and Konieczny, Lars and Hemforth, Barbara and Fedorenko, Evelina},
  journal={Psychological science},
  volume={28},
  number={6},
  pages={703--712},
  year={2017},
  publisher={Sage Publications Sage CA: Los Angeles, CA}
}

@article{ryskin2018comprehenders,
  title={Comprehenders model the nature of noise in the environment},
  author={Ryskin, Rachel and Futrell, Richard and Kiran, Swathi and Gibson, Edward},
  journal={Cognition},
  volume={181},
  pages={141--150},
  year={2018},
  publisher={Elsevier}
}

@article{chen2023effect,
  title={The effect of context on noisy-channel sentence comprehension},
  author={Chen, Sihan and Nathaniel, Sarah and Ryskin, Rachel and Gibson, Edward},
  journal={Cognition},
  volume={238},
  pages={105503},
  year={2023},
  publisher={Elsevier}
}

@article{ryskin2021erp,
  title={An ERP index of real-time error correction within a noisy-channel framework of human communication},
  author={Ryskin, Rachel and Stearns, Laura and Bergen, Leon and Eddy, Marianna and Fedorenko, Evelina and Gibson, Edward},
  journal={Neuropsychologia},
  volume={158},
  pages={107855},
  year={2021},
  publisher={Elsevier}
}

@article{tillman2020sequential,
  title={Sequential sampling models without random between-trial variability: The racing diffusion model of speeded decision making},
  author={Tillman, Gabriel and Van Zandt, Trish and Logan, Gordon D},
  journal={Psychonomic Bulletin \& Review},
  volume={27},
  number={5},
  pages={911--936},
  year={2020},
  publisher={Springer}
}

@article{simon1972theories,
  title={Theories of bounded rationality},
  author={Simon, Herbert A},
  journal={Decision and organization},
  volume={1},
  number={1},
  pages={161--176},
  year={1972},
  publisher={Amsterdam}
}

@article{altmann1998ambiguity,
  title={Ambiguity in sentence processing},
  author={Altmann, Gerry TM},
  journal={Trends in cognitive sciences},
  volume={2},
  number={4},
  pages={146--152},
  year={1998},
  publisher={Elsevier}
}

@Misc{loo,
    title = {loo: Efficient leave-one-out cross-validation and WAIC for
      Bayesian models},
    author = {Aki Vehtari and Jonah Gabry and Måns Magnusson and Yuling
      Yao and Paul-Christian Bürkner and Topi Paananen and Andrew
      Gelman},
    year = {2024},
    note = {R package version 2.8.0},
    url = {https://mc-stan.org/loo/},
  }

@article{radford2019language,
  title={Language Models are Unsupervised Multitask Learners},
  author={Radford, Alec and Wu, Jeff and Child, Rewon and Luan, David and Amodei, Dario and Sutskever, Ilya},
  year={2019}
}

@inproceedings{lu2024can,
  title={Can Syntactic Log-Odds Ratio Predict Acceptability and Satiation?},
  author={Lu, Jiayi and Merchan, Jonathan and Wang, Lian and Degen, Judith},
  booktitle={Proceedings of the Society for Computation in Linguistics 2024},
  pages={10--19},
  year={2024}
}

@article{oh2023does,
  title={Why does surprisal from larger transformer-based language models provide a poorer fit to human reading times?},
  author={Oh, Byung-Doh and Schuler, William},
  journal={Transactions of the Association for Computational Linguistics},
  volume={11},
  pages={336--350},
  year={2023},
  publisher={MIT Press One Broadway, 12th Floor, Cambridge, Massachusetts 02142, USA~…}
}

@article{shain2024large,
  title={Large-scale evidence for logarithmic effects of word predictability on reading time},
  author={Shain, Cory and Meister, Clara and Pimentel, Tiago and Cotterell, Ryan and Levy, Roger},
  journal={Proceedings of the National Academy of Sciences},
  volume={121},
  number={10},
  pages={e2307876121},
  year={2024},
  publisher={National Academy of Sciences}
}

@article{lau2020furiously,
  title={How furiously can colorless green ideas sleep? sentence acceptability in context},
  author={Lau, Jey Han and Armendariz, Carlos and Lappin, Shalom and Purver, Matthew and Shu, Chang},
  journal={Transactions of the Association for Computational Linguistics},
  volume={8},
  pages={296--310},
  year={2020},
  publisher={MIT Press One Rogers Street, Cambridge, MA 02142-1209, USA journals-info~…}
}

@inproceedings{levenshtein1966binary,
  title={Binary codes capable of correcting deletions, insertions, and reversals},
  author={Levenshtein, Vladimir I and others},
  booktitle={Soviet physics doklady},
  volume={10},
  number={8},
  pages={707--710},
  year={1966},
  organization={Soviet Union}
}

@article{damerau1964technique,
  title={A technique for computer detection and correction of spelling errors},
  author={Damerau, Fred J},
  journal={Communications of the ACM},
  volume={7},
  number={3},
  pages={171--176},
  year={1964},
  publisher={ACM New York, NY, USA}
}

@article{vehtari2017practical,
  title={Practical Bayesian model evaluation using leave-one-out cross-validation and WAIC},
  author={Vehtari, Aki and Gelman, Andrew and Gabry, Jonah},
  journal={Statistics and computing},
  volume={27},
  pages={1413--1432},
  year={2017},
  publisher={Springer}
}

@article{barr2013random,
  title={Random effects structure for confirmatory hypothesis testing: Keep it maximal},
  author={Barr, Dale J and Levy, Roger and Scheepers, Christoph and Tily, Harry J},
  journal={Journal of memory and language},
  volume={68},
  number={3},
  pages={255--278},
  year={2013},
  publisher={Elsevier}
}

@article{nalborczyk2019introduction,
  title={An introduction to Bayesian multilevel models using brms: A case study of gender effects on vowel variability in standard Indonesian},
  author={Nalborczyk, Ladislas and Batailler, C{\'e}dric and L{\oe}venbruck, H{\'e}l{\`e}ne and Vilain, Anne and B{\"u}rkner, Paul-Christian},
  journal={Journal of Speech, Language, and Hearing Research},
  volume={62},
  number={5},
  pages={1225--1242},
  year={2019},
  publisher={American Speech-Language-Hearing Association}
}

@article{r2024r,
  title={R: A language and environment for statistical computing},
  author={R Core Team, R and others},
  year={2024},
  publisher={R foundation for statistical computing Vienna, Austria}
}

@article{burkner2019ordinal,
  title={Ordinal regression models in psychology: A tutorial},
  author={B{\"u}rkner, Paul-Christian and Vuorre, Matti},
  journal={Advances in Methods and Practices in Psychological Science},
  volume={2},
  number={1},
  pages={77--101},
  year={2019},
  publisher={Sage Publications Sage CA: Los Angeles, CA}
}

@article{burkner2017brms,
  title={brms: An R package for Bayesian multilevel models using Stan},
  author={B{\"u}rkner, Paul-Christian},
  journal={Journal of statistical software},
  volume={80},
  pages={1--28},
  year={2017}
}

@article{poliak2024not,
  title={It is not what you say but how you say it: Evidence from Russian shows robust effects of the structural prior on noisy channel inferences.},
  author={Poliak, Moshe and Ryskin, Rachel and Braginsky, Mika and Gibson, Edward},
  journal={Journal of Experimental Psychology: Learning, Memory, and Cognition},
  volume={50},
  number={4},
  pages={637},
  year={2024},
  publisher={American Psychological Association}
}

@inproceedings{liu2020structural,
  title={Structural frequency effects in noisy-channel comprehension},
  author={Liu, Yingtong and Ryskin, Rachel and Futrell, Richard and Gibson, Edward},
  booktitle={Presentation at the Penn Linguistics Conference},
  year={2020}
}

@phdthesis{keller2000gradience,
  title={Gradience in grammar: Experimental and computational aspects of degrees of grammaticality},
  author={Keller, Frank},
  year={2000}
}

@article{hofmeister2012islands,
  title={Islands in the grammar? Standards of evidence},
  author={Hofmeister, Philip and Staum Casasanto, Laura and Sag, Ivan A},
  year={2012},
  publisher={Cambridge University Press}
}

@article{hofmeister2013source,
  title={The source ambiguity problem: Distinguishing the effects of grammar and processing on acceptability judgments},
  author={Hofmeister, Philip and Jaeger, T Florian and Arnon, Inbal and Sag, Ivan A and Snider, Neal},
  journal={Language and cognitive processes},
  volume={28},
  number={1-2},
  pages={48--87},
  year={2013},
  publisher={Taylor \& Francis}
}

@article{leivada2020acceptable,
  title={Acceptable ungrammatical sentences, unacceptable grammatical sentences, and the role of the cognitive parser},
  author={Leivada, Evelina and Westergaard, Marit},
  journal={Frontiers in psychology},
  volume={11},
  pages={364},
  year={2020},
  publisher={Frontiers Media SA}
}

@inproceedings{pauls2012large,
  title={Large-scale syntactic language modeling with treelets},
  author={Pauls, Adam and Klein, Dan},
  booktitle={Proceedings of the 50th Annual Meeting of the Association for Computational Linguistics (Volume 1: Long Papers)},
  pages={959--968},
  year={2012}
}

@article{hofmeister2010cognitive,
  title={Cognitive constraints and island effects},
  author={Hofmeister, Philip and Sag, Ivan A},
  journal={Language},
  volume={86},
  number={2},
  pages={366--415},
  year={2010},
  publisher={Linguistic Society of America}
}

@article{lau2017grammaticality,
  title={Grammaticality, acceptability, and probability: A probabilistic view of linguistic knowledge},
  author={Lau, Jey Han and Clark, Alexander and Lappin, Shalom},
  journal={Cognitive science},
  volume={41},
  number={5},
  pages={1202--1241},
  year={2017},
  publisher={Wiley Online Library}
}

@article{ferreira1991recovery,
  title={Recovery from misanalyses of garden-path sentences},
  author={Ferreira, Fernanda and Henderson, John M},
  journal={Journal of Memory and Language},
  volume={30},
  number={6},
  pages={725--745},
  year={1991},
  publisher={Elsevier}
}

@article{gibson1999memory,
  title={Memory limitations and structural forgetting: The perception of complex ungrammatical sentences as grammatical},
  author={Gibson, Edward and Thomas, James},
  journal={Language and Cognitive Processes},
  volume={14},
  number={3},
  pages={225--248},
  year={1999},
  publisher={Taylor \& Francis}
}

@book{lu2024linguistic,
  title={Linguistic Adaptation to Unacceptable Sentences},
  author={Lu, Jiayi},
  year={2024},
  publisher={Stanford University}
}

@book{schutze1996empirical,
  title={The empirical base of linguistics: Grammaticality judgments and linguistic methodology},
  author={Sch{\"u}tze, Carson},
  year={1996},
  publisher={University of Chicago Press}
}

@article{tonhauser2015empirical,
  title={Empirical evidence in research on meaning},
  author={Tonhauser, Judith and Matthewson, Lisa},
  journal={Ms., The Ohio State University and University of British Columbia},
  year={2015},
  publisher={Citeseer}
}

@book{kelley2018more,
  title={More People Understand Eschers than the Linguist Does: The Causes and Effects of Grammatical Illusions},
  author={Kelley, Patrick},
  year={2018},
  publisher={Michigan State University}
}

@book{townsend2001sentence,
  title={Sentence comprehension: The integration of habits and rules},
  author={Townsend, David J and Bever, Thomas G},
  year={2001},
  publisher={MIT Press}
}

@article{paape2024linguistic,
  title={How do linguistic illusions arise? Rational inference and good-enough processing as competing latent processes within individuals},
  author={Paape, Dario},
  journal={Language, Cognition and Neuroscience},
  volume={39},
  number={10},
  pages={1334--1365},
  year={2024},
  publisher={Taylor \& Francis}
}

@incollection{christensen2016dead,
  title={The dead ends of language: The (mis) interpretation of a grammatical illusion},
  author={Christensen, Ken Ramsh{\o}j},
  booktitle={Let us have articles betwixt us: Papers in Historical and Comparative Linguistics in Honour of Johanna L. Wood},
  pages={129--159},
  year={2016},
  publisher={Department of English, University of Aarhus}
}

@article{de2016more,
  title={More people have presented in conferences than I have. Comparative illusions: When ungrammaticality goes unnoticed},
  author={De Dios-Flores, Iria},
  journal={Glancing Backwards To Build a Future in English Studies},
  pages={219},
  year={2016}
}

@inproceedings{oconnor2013,
  title={Evidence for online repair of Escher sentences},
  author={O'Connor, Ellen and Pancheva, Roumyana and Kaiser, Elsi},
  booktitle={Proceedings of Sinn und Bedeutung},
  volume={17},
  pages={363--380},
  year={2013}
}

@misc{liberman2018,
  author       = {Mark Liberman},
  title        = {Escher sentences},
  year         = {2018},
  url          = {http://itre.cis.upenn.edu/~myl/languagelog/archives/000862.html},
  note         = {Accessed: 2025-03-18},
  howpublished = {Language Log}
}

@misc{pullum2004,
  author       = {Geoffrey K. Pullum},
  title        = {Plausible Angloid gibberish},
  year         = {2004},
  url          = {http://itre.cis.upenn.edu/~myl/languagelog/archives/000860.html},
  note         = {Accessed: 2025-03-18},
  howpublished = {Language Log}
}

@misc{pullum2009,
  author       = {Geoffrey K. Pullum},
  title        = {More people than you think will understand},
  year         = {2009},
  month        = {12},
  url          = {https://languagelog.ldc.upenn.edu/nll/?p=1997},
  note         = {Accessed: 2025-03-18},
  howpublished = {Language Log}
}

@phdthesis{montalbetti1984after,
  title={After binding: On the interpretation of pronouns},
  author={Montalbetti, Mario M},
  year={1984},
  school={Massachusetts Institute of Technology}
}

@article{snyder2000experimental,
  title={An experimental investigation of syntactic satiation effects},
  author={Snyder, William},
  journal={Linguistic inquiry},
  volume={31},
  number={3},
  pages={575--582},
  year={2000},
  publisher={JSTOR}
}

@article{phillips2011grammatical,
  title={Grammatical illusions and selective fallibility in real-time language comprehension},
  author={Phillips, Colin and Wagers, Matthew W and Lau, Ellen F},
  year={2011},
  journal={Experiments at the Interfaces},
  volume={37},
  pages={147--180},
}

@article{hahn2022resource,
  title={A resource-rational model of human processing of recursive linguistic structure},
  author={Hahn, Michael and Futrell, Richard and Levy, Roger and Gibson, Edward},
  journal={Proceedings of the National Academy of Sciences},
  volume={119},
  number={43},
  pages={e2122602119},
  year={2022},
  publisher={National Academy of Sciences}
}

@article{levy2008expectation,
  title={Expectation-based syntactic comprehension},
  author={Levy, Roger},
  journal={Cognition},
  volume={106},
  number={3},
  pages={1126--1177},
  year={2008},
  publisher={Elsevier}
}

@inproceedings{poliak2025,
	address = {Saarbruecken, Germany},
	title = {Rational Inference Underlies Judgments of Grammatical Well-Formedness},
	url = {https://nds.uni-saarland.de/rails2025/submission_22.pdf},
	language = {en},
	urldate = {2025},
	booktitle = {Conference on {Rational} {Approaches} in {Language} {Science} (RAILS)},
	author = {Poliak, Moshe and An, Aixiu and Levy, Roger and Gibson, Edward},
	year = {2025},
}

@article{misra2022minicons,
  title={minicons: Enabling flexible behavioral and representational analyses of transformer language models},
  author={Misra, Kanishka},
  journal={arXiv preprint arXiv:2203.13112},
  year={2022}
}

@phdthesis{oconnor2015comparative,
  title={Comparative iilusions at the syntax-semantics interface},
  author={O'Connor, Ellen},
  year={2015},
  school={University of Southern California}
}

@article{wellwood2018anatomy,
  title={The anatomy of a comparative illusion},
  author={Wellwood, Alexis and Pancheva, Roumyana and Hacquard, Valentine and Phillips, Colin},
  journal={Journal of Semantics},
  volume={35},
  number={3},
  pages={543--583},
  year={2018},
  publisher={Oxford University Press}
}

@inproceedings{levy_noisy-channel_2008,
	address = {Honolulu, Hawaii},
	title = {A noisy-channel model of rational human sentence comprehension under uncertain input},
	url = {http://portal.acm.org/citation.cfm?doid=1613715.1613749},
	doi = {10.3115/1613715.1613749},
	language = {en},
	urldate = {2020-02-12},
	booktitle = {Proceedings of the {Conference} on {Empirical} {Methods} in {Natural} {Language} {Processing} - {EMNLP} '08},
	publisher = {Association for Computational Linguistics},
	author = {Levy, Roger},
	year = {2008},
	pages = {234},
}

@article{gibson_rational_2013,
	title = {Rational integration of noisy evidence and prior semantic expectations in sentence interpretation},
	volume = {110},
	url = {https://www.pnas.org/doi/abs/10.1073/pnas.1216438110},
	doi = {10.1073/pnas.1216438110},
	number = {20},
	urldate = {2023-10-26},
	journal = {Proceedings of the National Academy of Sciences},
	author = {Gibson, Edward and Bergen, Leon and Piantadosi, Steven T.},
	month = may,
	year = {2013},
	note = {Publisher: Proceedings of the National Academy of Sciences},
	pages = {8051--8056},
}

@article{meylan2023adults,
  title={How adults understand what young children say},
  author={Meylan, Stephan C and Foushee, Ruthe and Wong, Nicole H and Bergelson, Elika and Levy, Roger P},
  journal={Nature Human Behaviour},
  volume={7},
  number={12},
  pages={2111--2125},
  year={2023},
  publisher={Nature Publishing Group UK London}
}

@article{paape_quadruplex_2020,
	title = {Quadruplex negatio invertit? {The} on-line processing of depth charge sentences},
	volume = {37},
	copyright = {http://creativecommons.org/licenses/by/4.0/},
	issn = {0167-5133, 1477-4593},
	shorttitle = {Quadruplex {Negatio} {Invertit}?},
	language = {en},
	number = {4},
	urldate = {2024-05-16},
	journal = {Journal of Semantics},
	author = {Paape, Dario and Vasishth, Shravan and Von Der Malsburg, Titus},
	month = nov,
	year = {2020},
	pages = {509--555},
}

@article{zhang_comparative_2024,
	title = {Comparative illusions are evidence of rational inference in language comprehension},
	copyright = {https://creativecommons.org/publicdomain/zero/1.0/legalcode},
	author = {Zhang, Yuhan and Kauf, Carina and Levy, Roger Philip and Gibson, Edward},
	year = {2025},
        journal = {Journal of Experimental Psychology: General},
        doi = {10.1037/xge0001807},
}

@article{dahan2019language,
  title={Language comprehension: Insights from research on spoken language},
  author={Dahan, Delphine and Ferreira, Fernanda},
  journal={Human language: from genes and brains to behavior},
  pages={21--33},
  year={2019},
  publisher={MIT Press}
}

@article{levelt1983monitoring,
  title={Monitoring and self-repair in speech},
  author={Levelt, Willem JM},
  journal={Cognition},
  volume={14},
  number={1},
  pages={41--104},
  year={1983},
  publisher={Elsevier}
}

@misc{zhang_memory-based_2024,
      title={Noisy memory encoding explains negative polarity illusions}, 
      author={Yuhan Zhang and Edward Gibson},
      year={2026},
      eprint={2606.04340},
      archivePrefix={arXiv},
      primaryClass={cs.CL},
      url={https://arxiv.org/abs/2606.04340}, 
}

@article{christianson2006younger,
  title={Younger and older adults'" good-enough" interpretations of garden-path sentences},
  author={Christianson, Kiel and Williams, Carrick C and Zacks, Rose T and Ferreira, Fernanda},
  journal={Discourse processes},
  volume={42},
  number={2},
  pages={205--238},
  year={2006},
  publisher={Taylor \& Francis}
}

@article{zhang_noisy-channel_2023,
	title = {A noisy-channel approach to depth-charge illusions},
	volume = {232},
	issn = {00100277},
	language = {en},
	urldate = {2022-12-12},
	journal = {Cognition},
	author = {Zhang, Yuhan and Ryskin, Rachel and Gibson, Edward},
	month = mar,
	year = {2023},
	pages = {105346},
}

@article{ryskin_agreement_2021,
    title = {Agreement errors are predicted by rational inference in sentence processing},
    url = {https://psyarxiv.com/uaxsq/download?format=pdf},
    urldate = {2024-02-04},
    author = {Ryskin, Rachel and Bergen, Leon and Gibson, Edward},
    year = {2021},
    note = {Publisher: PsyArXiv},
}

@inproceedings{zhang_can_2023,
    title = {Can {Language} {Models} {Be} {Tricked} by {Language} {Illusions}? {Easier} with {Syntax}, {Harder} with {Semantics}},
    shorttitle = {Can {Language} {Models} {Be} {Tricked} by {Language} {Illusions}?},
    url = {http://arxiv.org/abs/2311.01386},
    doi = {10.18653/v1/2023.conll-1.1},
    urldate = {2024-04-02},
    booktitle = {Proceedings of the 27th {Conference} on {Computational} {Natural} {Language} {Learning} ({CoNLL})},
    author = {Zhang, Yuhan and Gibson, Edward and Davis, Forrest},
    year = {2023},
    note = {arXiv:2311.01386 [cs]},
    pages = {1--14},
}

@article{christianson_when_2016,
    title = {When language comprehension goes wrong for the right reasons: {Good}-enough, underspecified, or shallow language processing},
    volume = {69},
    issn = {1747-0218},
    shorttitle = {When language comprehension goes wrong for the right reasons},
    url = {https://doi.org/10.1080/17470218.2015.1134603},
    doi = {10.1080/17470218.2015.1134603},
    language = {EN},
    number = {5},
    urldate = {2026-05-14},
    journal = {Quarterly Journal of Experimental Psychology},
    publisher = {SAGE Publications},
    author = {Christianson, Kiel},
    month = may,
    year = {2016},
    pages = {817--828},
}

@article{karimi_good-enough_2016,
    title = {Good-enough linguistic representations and online cognitive equilibrium in language processing},
    volume = {69},
    issn = {1747-0218},
    url = {https://doi.org/10.1080/17470218.2015.1053951},
    doi = {10.1080/17470218.2015.1053951},
    language = {EN},
    number = {5},
    urldate = {2026-05-14},
    journal = {Quarterly Journal of Experimental Psychology},
    publisher = {SAGE Publications},
    author = {Karimi, Hossein and Ferreira, Fernanda},
    month = may,
    year = {2016},
    pages = {1013--1040},
}

@incollection{chafe_integration_1982,
    title = {Integration and involvement in speaking, writing, and oral literature},
    booktitle = {Spoken and {Written} {Language}: {Exploring} {Orality} and {Literacy}},
    publisher = {Ablex},
    author = {Chafe, Wallace},
    editor = {Tannen, D.},
    year = {1982},
}

@incollection{garrett_analysis_1975,
    title = {The {Analysis} of {Sentence} {Production}},
    volume = {9},
    isbn = {978-0-12-543309-9},
    url = {https://linkinghub.elsevier.com/retrieve/pii/S0079742108602704},
    doi = {10.1016/S0079-7421(08)60270-4},
    language = {en},
    urldate = {2024-11-11},
    booktitle = {Psychology of {Learning} and {Motivation}},
    publisher = {Elsevier},
    author = {Garrett, M.F.},
    year = {1975},
    pages = {133--177},
}

\appendix
\setcounter{table}{0} 
\renewcommand{\thetable}{A\arabic{table}} 

\setcounter{figure}{0} 
\renewcommand{\thefigure}{A\arabic{figure}} 

\section{Appendix}
\subsection{Experiment 1: Statistical analysis}

Table \ref{tab:appendix-model-comparison-exp1} shows the performance of Bayesian multilevel ordinal regression models that differ by the random-effects structure. Lower LOOIC (leave-one-out information criterion) indicates better performance. Table \ref{tab:appendix-model-comparison-exp1} shows that when the random effects took the full structure of the fixed effects, the model performed better. 

\begin{table}[h]
    \centering
    \begin{tabular}{p{0.1\linewidth}p{0.1\linewidth}p{0.1\linewidth}p{0.1\linewidth}p{0.1\linewidth}p{0.4\linewidth}}\toprule
    Model  & LOOIC & SE & $\Delta$LOOIC & $\Delta$SE & Right side of the regression model \\ \midrule
    Exp1 & 39031.3 & 238.1 & 0.00 & 0.00 & sub-form * sub-number + (1 + sub-form * sub-number|participant) + (1 + sub-form * sub-number|item) \\ \midrule
    base & 41323.0 & 230.6 & 2291.7 & -7.5 & sub-form * sub-number + (1|participant) + (1|item) \\ \bottomrule
    \end{tabular}
    \caption{Model comparison with LOOIC in Experiment 1 (smaller LOOIC indicates better predictive performance)}
    \label{tab:appendix-model-comparison-exp1}
\end{table}

Table \ref{tab:appendic-stat-exp1} represents the posterior estimates for the coefficients for the regression model reported in the main paper.

\begin{table}[h]
\centering
\begin{tabular}{p{0.3\textwidth}p{0.1\textwidth}p{0.1\textwidth}p{0.08\textwidth}p{0.08\textwidth}p{0.06\textwidth}p{0.1\textwidth}}
\hline
\textbf{Parameter} & \textbf{Estimate} & \textbf{Est.Error} & \textbf{l-95\% CI} & \textbf{u-95\% CI} & \textbf{Rhat} & \textbf{Bulk ESS} \\
\hline
Intercept[1]                  & -5.28 & 0.17 & -5.62 & -4.94 & 1.00 & 1462  \\
Intercept[2]                  & -3.55 & 0.17 & -3.87 & -3.22 & 1.00 & 1379  \\
Intercept[3]                  & -2.36 & 0.16 & -2.68 & -2.04 & 1.00 & 1354  \\
Intercept[4]                  & -1.98 & 0.16 & -2.30 & -1.65 & 1.00 & 1354  \\
Intercept[5]                  & -0.82 & 0.16 & -1.14 & -0.50 & 1.00 & 1339  \\
Intercept[6]                  &  0.98 & 0.16 &  0.66 &  1.30 & 1.00 & 1350  \\
subjectnp                     & -1.95 & 0.15 & -2.25 & -1.65 & 1.00 & 3225  \\
conditionplural               & -0.28 & 0.07 & -0.43 & -0.14 & 1.00 & 10937  \\
conditioncontrol              &  0.02 & 0.16 & -0.31 &  0.35 & 1.00 & 4170  \\
subjectnp:conditionplural     &  2.14 & 0.19 &  1.77 &  2.51 & 1.00 & 4725  \\
subjectnp:conditioncontrol    &  2.72 & 0.25 &  2.24 &  3.20 & 1.00 & 4481  \\
\hline
\end{tabular}
\caption{Posterior estimates for the reported statistical model in Experiment 1}
\label{tab:appendic-stat-exp1}
\end{table}

\newpage
\subsection{Exploration Phase: Selecting the posterior predictors from published data} \label{sec:appendix_exploratory_phase}

This section reviews the exploratory phase results where the goal was to predict the acceptability of CI sentences from the published dataset in \textcite{zhang_comparative_2024}.

\subsubsection{GPT-2 Small as LM}

\begin{figure}[htbp]
\centering
\includegraphics[width=0.95\linewidth]{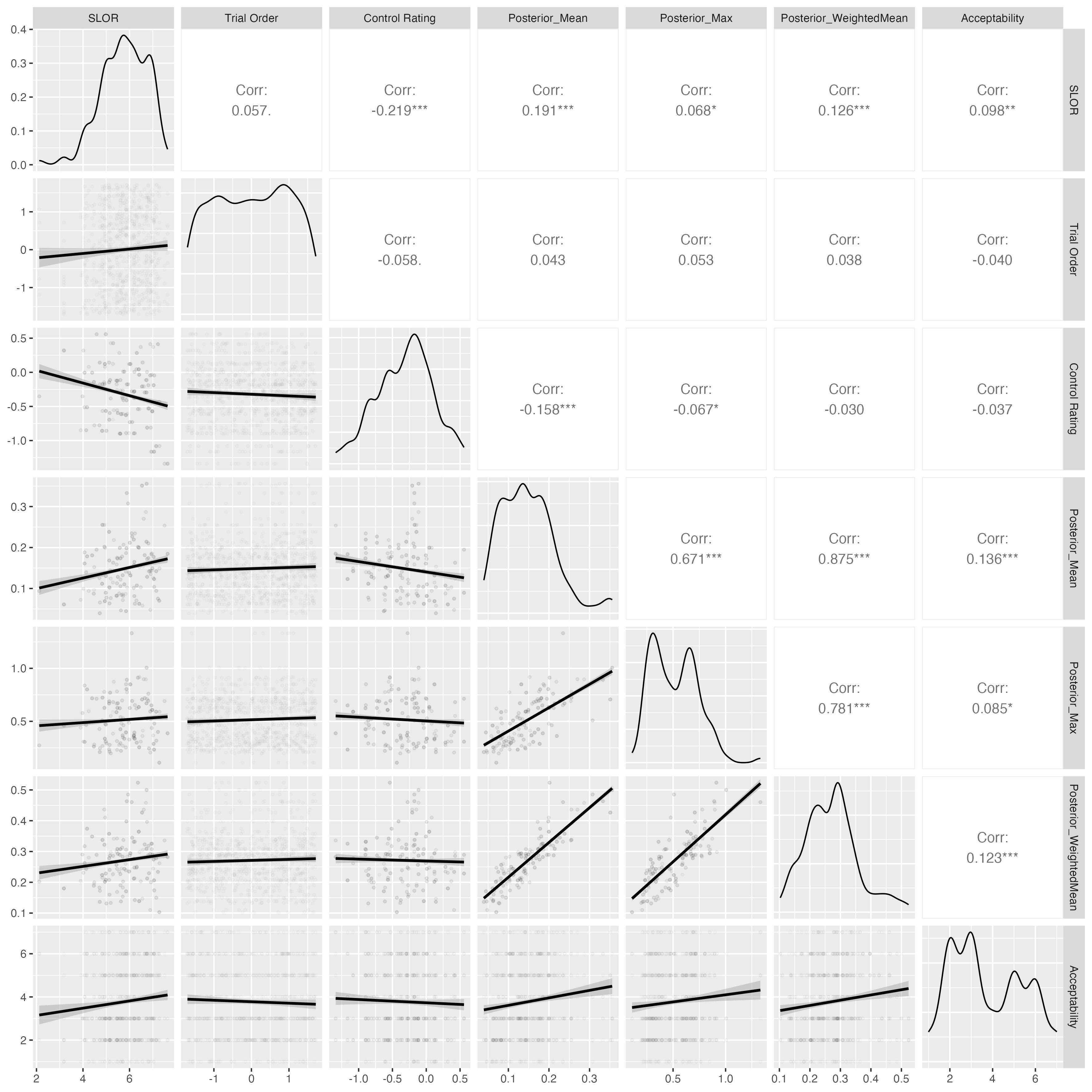}
\caption{\textbf{GPT-2 Small \& $p(s)$ is LM derived}: Pairwise correlation matrix for acceptability, SLOR, presentation order, acceptability of the control, and the posterior metrics (The diagonal line represents the density plot for each metric; the ``Corr'' cells represent the Pearson correlation coefficient and the significance value; the cells in the bottom-left triangle represent scatter plots of the two compared variables and a linear regression line with shadings representing the standard error.)}
\label{fig:corr-measures-gpt2-hack-old-data}
\end{figure}

As an exploratory analysis, we plotted the pairwise correlation between the response variable and each of the explanatory variables in Figure \ref{fig:corr-measures-gpt2-hack-old-data} where the probabilities of sentences were taken from GPT-2 Small and the encoding of $p(s)$ was directly derived from GPT-2 Small. The acceptability score is positively correlated with SLOR and all of the three aggregate measures of the posterior probability of the plausible. Since there is multicollinearity among SLOR and the other three posteriors, the statistical modeling should reveal whether each of those variables can independently predict acceptability rating while the others are controlled.

For the statistical modeling, we ran eight Bayesian multilevel cumulative ordinal regression models with varying explanatory variables as the population-level effects to predict the acceptability of illusory materials. The model design is in Table \ref{tab:structure-stat-model}. The acceptability score is on a likert scale from 1 to 7 and the models assume a non-equidistant interval between points on the scale. All explanatory variables were standardized to ensure they had a grand mean of zero. The group-level effect included random intercepts for different participants and remained the same across all models. We set weakly informative priors for all models. Both the intercepts and the population-level coefficients had a prior Gaussian distribution ($\text{Normal}(0,2)$) and the by-participant varying intercepts were given a weakly informative Half-Cauchy prior ($\text{Cauchy}(0,10)$), following the set up in Experiment 1. Four sampling chains were initiated, with 2000 iterations each and 1000 iterations as the warmup. The sampling chains for all coefficients in all models had a $\hat{R}$ of 1, indicating successful convergence.

We relied on the \texttt{loo} package \parencite{loo} in R to conduct the Bayesian leave-one-out-cross-validation \parencite[LOO-CV][]{vehtari2017practical} to obtain an approximation of the models' predictive abilities. Table \ref{tab:looic_gpt2_hack_old_data} shows the performance of eight statistical models with GPT-2 probability measures. LOOIC represents the deviance between the distribution of predictions and the distribution of true observables and a lower LOOIC of a model represents its better predictive performance. It is not hard to tell that out of the three aggregate posterior metrics, the \textsc{mean} metric outperforms the other two mathematical aggregates of the posterior probability of plausible alternatives, when only one metric was selected.

\begin{table}[htbp]
\centering
\begin{tabular}{lrrrrrr}
\toprule
Model & elpd\_diff & se\_diff & elpd\_loo & se\_elpd\_loo & looic & se\_looic \\
\midrule
mean             &  0.00 &  0.00 & -1334.98 & 23.35 & 2669.96 & 46.70  \\
mean, weightMean & -1.15 &  0.03 & -1336.13 & 23.38 & 2672.26 & 46.77  \\
mean, max        & -1.18 &  0.04 & -1336.16 & 23.39 & 2672.31 & 46.77  \\
weightMean       & -1.58 &  0.06 & -1336.56 & 23.41 & 2673.11 & 46.81  \\
all three       & -1.97 &  0.05 & -1336.95 & 23.39 & 2673.89 & 46.79  \\
max, weightMean  & -2.07 &  0.08 & -1337.05 & 23.43 & 2674.10 & 46.86 \\
max              & -4.72 &  0.00 & -1339.70 & 23.35 & 2679.39 & 46.70 \\
base             & -5.01 & -0.01 & -1339.99 & 23.34 & 2679.99 & 46.67 \\
\bottomrule
\end{tabular}
\caption{(\textbf{GPT-2 Small, LM-derived $p(s)$)}) Model comparison based on ELPD and LOOIC with associated standard errors}
\label{tab:looic_gpt2_hack_old_data}
\end{table}

Table \ref{tab:coefficient_gpt2_hack_old_data} reveals the estimate coefficients of the explanatory variables for eight statistical models. The SLOR variable positively predicts the acceptability score across all eight models. The trial order and the acceptability of the control sentence do not effectively impact the acceptability of CI sentences. When entered as the sole posterior predictor, the \textsc{mean} and the \textsc{weighted mean} posterior predictors positively predict the response variable, contrary to the \textsc{max} posterior predictor which has a 95\% probability including zero. 

\begin{table}[htbp]
\centering
\small
\setlength{\tabcolsep}{4pt}
\setlength{\extrarowheight}{2pt} 
\begin{adjustbox}{width=\linewidth}
\begin{tabular}{lcccccc}
\hline
\textbf{model} & \textbf{SLOR} & \textbf{Trial Order} & \textbf{Control Accept} &
\textbf{Posterior(max)} & \textbf{Posterior(mean)} & \textbf{Posterior(weighted mean)} \\ \hline
mean             & $0.180\,[0.042,\,0.318]$ & $-0.096\;[-0.222,\,0.032]$ & $0.063\;[-0.280,\,0.411]$ & --- & $0.257\,[0.116,\,0.401]$ & --- \\
mean, weightMean & $0.180\,[0.043,\,0.318]$ & $-0.097\;[-0.225,\,0.031]$ & $0.063\;[-0.288,\,0.411]$ & --- & $0.273\;[-0.034,\,0.573]$ & $-0.017\;[-0.300,\,0.269]$ \\
mean, max        & $0.178\,[0.045,\,0.313]$ & $-0.095\;[-0.221,\,0.032]$ & $0.069\;[-0.279,\,0.418]$ & $-0.042\;[-0.217,\,0.130]$ & $0.285\,[0.107,\,0.465]$ & --- \\
weightMean       & $0.184\,[0.049,\,0.321]$ & $-0.101\;[-0.232,\,0.027]$ & $0.040\;[-0.308,\,0.388]$ & --- & --- & $0.209\,[0.079,\,0.341]$ \\
all three       & $0.177\,[0.043,\,0.311]$ & $-0.097\;[-0.225,\,0.031]$ & $0.062\;[-0.282,\,0.405]$ & $-0.052\;[-0.267,\,0.160]$ & $0.264\;[-0.035,\,0.572]$ & $0.028\;[-0.317,\,0.376]$ \\
max, weightMean  & $0.178\,[0.040,\,0.316]$ & $-0.100\;[-0.227,\,0.026]$ & $0.035\;[-0.314,\,0.374]$ & $-0.084\;[-0.296,\,0.127]$ & --- & $0.273\,[0.067,\,0.482]$ \\
max              & $0.197\,[0.062,\,0.332]$ & $-0.101\;[-0.229,\,0.029]$ & $0.081\;[-0.257,\,0.426]$ & $0.131\;[-0.005,\,0.268]$ & --- & --- \\
base             & $0.191\,[0.053,\,0.328]$ & $-0.095\;[-0.218,\,0.031]$ & $0.115\;[-0.228,\,0.456]$ & --- & --- & --- \\
\hline
\end{tabular} 
\end{adjustbox}
\caption{\textbf{GPT-2 Small}: Estimate coefficients of statistical models (The names in the first column refer to which posterior variable(s) has/have been entered into the statistical model; For each column from the second, the numbers represent mean [95\% Credible Interval].)} \label{tab:coefficient_gpt2_hack_old_data}
\end{table}

\subsubsection{OPT as LM}

\begin{figure}[htbp]
\centering
\includegraphics[width=0.95\linewidth]{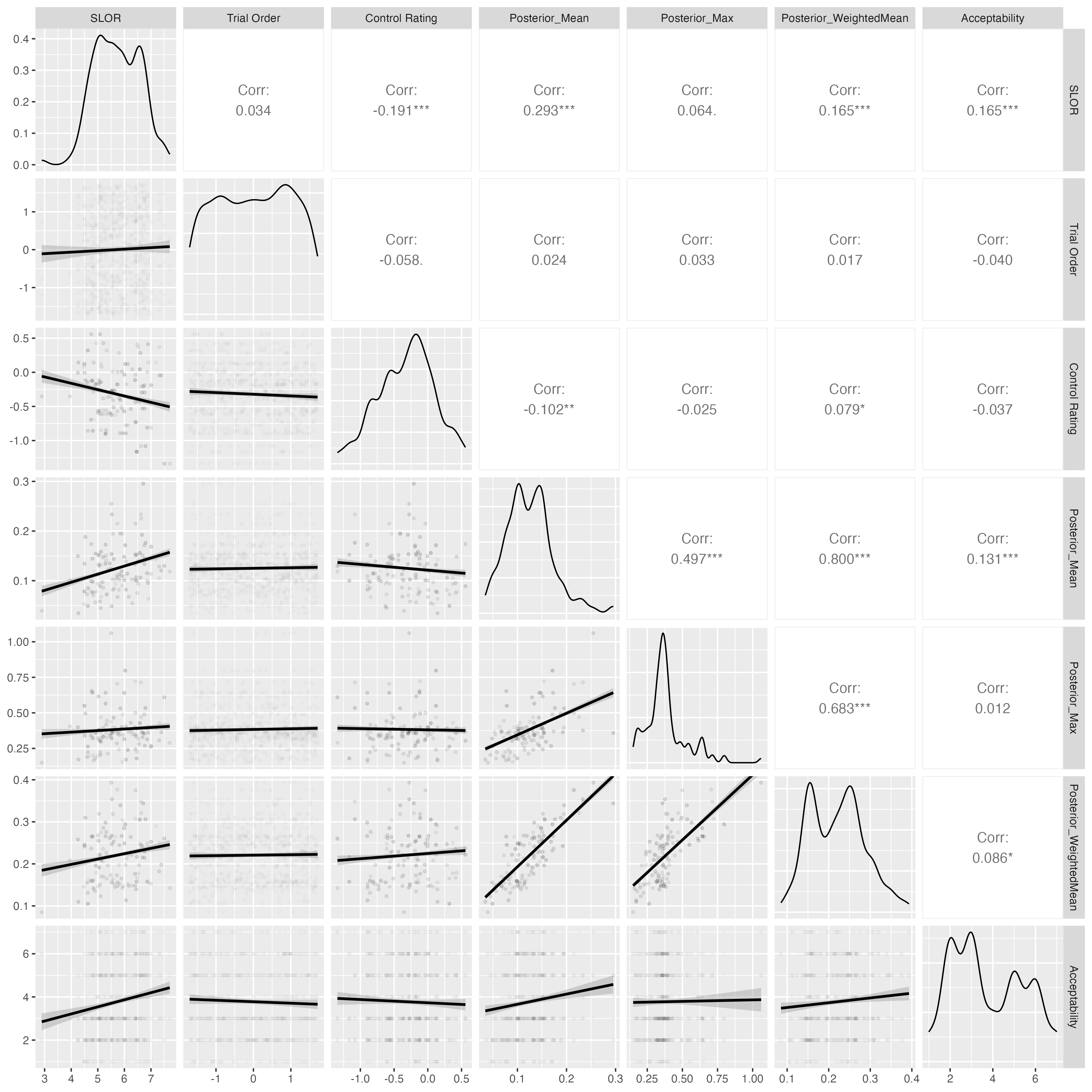}
\caption{\textbf{OPT \& $p(s)$ is LM derived}: Pairwise correlation matrix for acceptability, SLOR, presentation order, acceptability of the control, and the posterior metrics}
\label{fig:corr-measures-opt-hack-old-data}
\end{figure}

In a similar vein, we replaced GPT-2 Small with OPT to conduct the same analysis. Figure \ref{fig:corr-measures-opt-hack-old-data} replicates the correlation patterns in Figure \ref{fig:corr-measures-gpt2-hack-old-data}, except for the correlation between CI acceptability and the \textsc{max} posterior variable. With OPT, there isn't a significant positive correlation.

Table \ref{tab:looic_opt_hack_old_data} shows the performance of the eight statistical models with OPT probability measures. When entered as the sole posterior predictor, \textsc{mean} and \textsc{weighted mean} outperform \textsc{max}. Adding \textsc{max} into the baseline model with control variables even slightly decreases the model's predictive performance (see the last and the second to last rows in Table \ref{tab:looic_opt_hack_old_data}). 

\begin{table}[htbp]
\centering
\begin{tabular}{lrrrrrr}
\toprule
Model & elpd\_diff & se\_diff & elpd\_loo & se\_elpd\_loo & looic & se\_looic \\
\midrule
mean, max        &  0.00 &  0.00 & -1326.06 & 23.71 & 2652.12 & 47.42  \\
mean             & -0.59 & -0.01 & -1326.65 & 23.69 & 2653.31 & 47.39  \\
max, weightMean  & -0.70 &  0.02 & -1326.76 & 23.73 & 2653.52 & 47.46  \\
mean, weightMean & -1.21 & -0.01 & -1327.27 & 23.70 & 2654.53 & 47.40  \\
all three       & -1.26 &  0.00 & -1327.32 & 23.71 & 2654.64 & 47.43  \\
weightMean       & -2.31 & -0.01 & -1328.37 & 23.70 & 2656.73 & 47.39   \\
base             & -3.84 & -0.07 & -1329.90 & 23.64 & 2659.81 & 47.28   \\
max              & -4.31 & -0.04 & -1330.37 & 23.67 & 2660.74 & 47.35  \\
\bottomrule
\end{tabular}
\caption{(\textbf{OPT, LM-derived $p(s)$}) Model comparison with ELPD, LOOIC, and related metrics}
\label{tab:looic_opt_hack_old_data}
\end{table}

Table \ref{tab:coefficients_opt_hack_old_data} shows the estimate coefficients of the predictors for eight statistical models. Similar patterns emerge from the OPT route: when only one posterior variable is entered into the model, \textsc{mean} and \textsc{weighted mean} outperform \textsc{max}. More interestingly, the coefficient of \textsc{max} posterior decreases when added alongside either of the mean-related variable, suggesting when the central tendency of the posterior distribution of plausible alternatives is fixed, a larger posterior probability on the edge drives down the acceptability of CI sentences. This observation implies that during CI comprehension, comprehenders track the central tendency of the plausible alternative distributions, instead of being attacted to the most probable alternative. 

\begin{table}[htbp]
\centering
\small                    
\begin{adjustbox}{width=\linewidth}
\setlength{\tabcolsep}{4pt}   
\setlength{\extrarowheight}{2pt} 
\begin{tabular}{lcccccc}
\hline
\textbf{model} & \textbf{SLOR} & \textbf{Trial Order} & \textbf{Control Accept} &
\textbf{Posterior(max)} & \textbf{Posterior(mean)} & \textbf{Posterior(weighted mean)} \\ \hline
mean, max        & $0.314\,[0.170,\,0.455]$ & $-0.092\;[-0.222,\,0.036]$ & $0.077\;[-0.260,\,0.420]$ & $-0.122\;[-0.267,\,0.023]$ & $0.274\,[0.112,\,0.434]$ & --- \\
mean             & $0.326\,[0.184,\,0.468]$ & $-0.095\;[-0.227,\,0.034]$ & $0.079\;[-0.255,\,0.422]$ & ---                           & $0.204\,[0.066,\,0.344]$ & --- \\
max, weightMean  & $0.318\,[0.173,\,0.460]$ & $-0.096\;[-0.225,\,0.033]$ & $0.061\;[-0.278,\,0.401]$ & $-0.186\;[-0.359,\,-0.013]$ & ---                       & $0.280\,[0.096,\,0.460]$ \\
mean, weightMean & $0.327\,[0.185,\,0.470]$ & $-0.094\;[-0.225,\,0.036]$ & $0.087\;[-0.265,\,0.436]$ & ---                           & $0.282\,[0.009,\,0.556]$ & $-0.087\;[-0.347,\,0.171]$ \\
all three       & $0.311\,[0.168,\,0.450]$ & $-0.093\;[-0.221,\,0.036]$ & $0.073\;[-0.275,\,0.415]$ & $-0.146\;[-0.327,\,0.040]$  & $0.225\;[-0.052,\,0.503]$ & $0.069\;[-0.253,\,0.392]$ \\
weightMean       & $0.340\,[0.198,\,0.482]$ & $-0.098\;[-0.225,\,0.029]$ & $0.081\;[-0.262,\,0.424]$ & ---                           & ---                       & $0.140\,[0.008,\,0.273]$ \\
base             & $0.367\,[0.229,\,0.508]$ & $-0.095\;[-0.225,\,0.033]$ & $0.120\;[-0.224,\,0.453]$ & ---                           & ---                       & --- \\
max              & $0.367\,[0.227,\,0.506]$ & $-0.095\;[-0.222,\,0.033]$ & $0.119\;[-0.213,\,0.459]$ & $-0.000\;[-0.123,\,0.122]$   & ---                       & --- \\ \hline
\end{tabular}
\end{adjustbox}
\caption{\textbf{OPT}: Estimate coefficients of statistical models (The names in the first column refer to which posterior variable(s) has/have been entered into the statistical model; For each column from the second, the numbers represent mean [95\% Credible Interval].)} \label{tab:coefficients_opt_hack_old_data}
\end{table}

The statistical patterns in the exploratory phase show that the \textsc{mean} and \textsc{weighted mean} posterior variables outperformed the \textsc{max} posterior, which is generalizable in both language models. Since the \textsc{weighted mean} posterior variable can be regarded as a variant of \textsc{mean}, we only report \textsc{mean} in comparison to \textsc{max} in the generalization phase in the main paper.

\newpage
\subsection{Generalization Phase: Confirmatory results from OPT} \label{sec:appendix_opt_generalization}

\begin{figure}[htbp]
\centering
\includegraphics[width=0.95\linewidth]{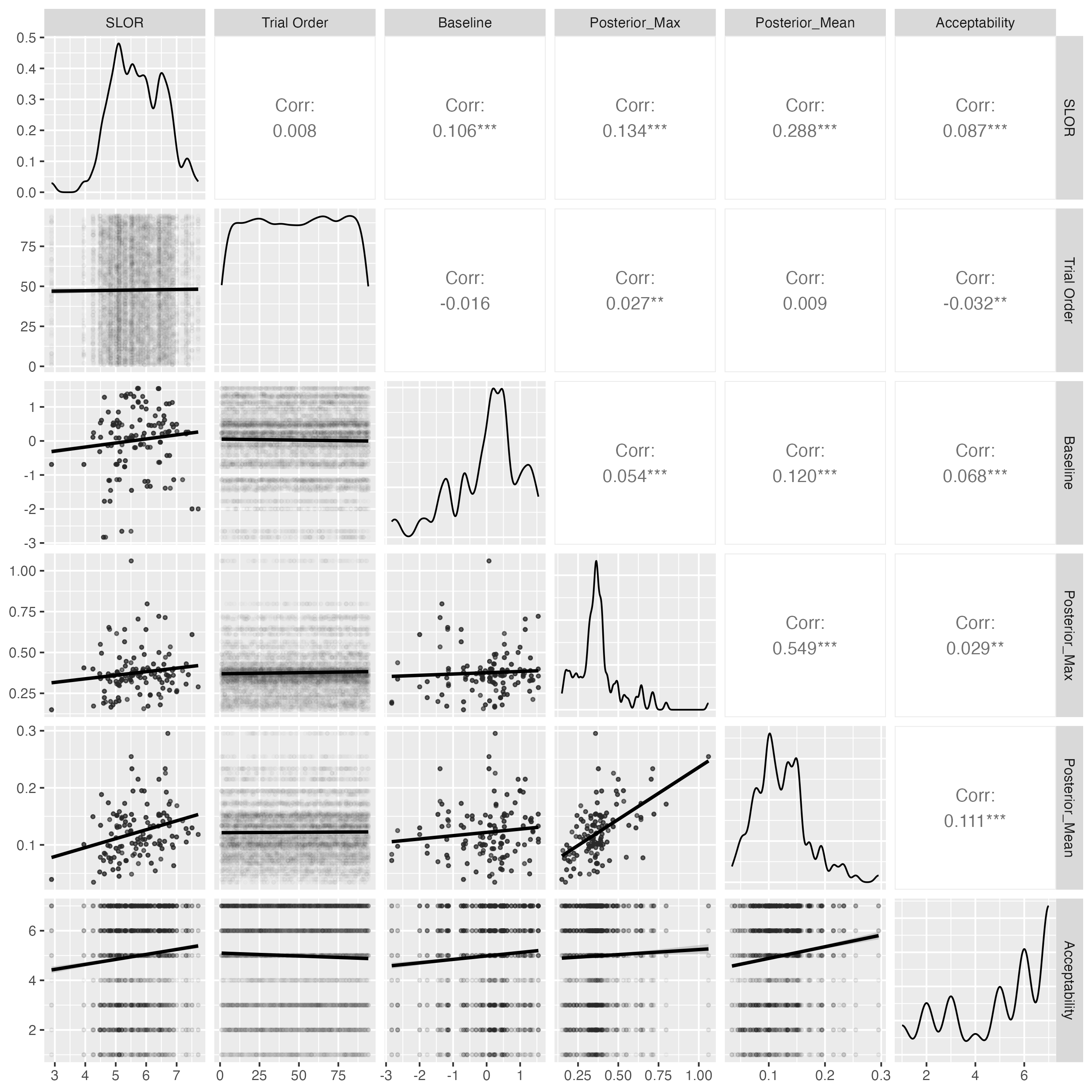}
\caption{\textbf{OPT}: Pairwise correlation matrix for acceptability, SLOR, trial order, acceptability of the baseline control sentence, and two posterior metrics (The diagonal line represents the density plot for each metric; the ``Corr'' cells represent the Pearson correlation coefficient and the significance value; the cells in the bottom-left triangle represent scatter plots of the two compared variables and a linear regression line with shadings representing the standard error. All variables except for the control sentence rating are on their original scale.)}
\label{fig:corr-measures-opt-hack-new-data}
\end{figure}

We conducted the same analysis for the generalization dataset with OPT as the underlying language model. The same statistical patterns emerge with this new LM. Figure \ref{fig:corr-measures-opt-hack-new-data} reveals the positive correlations between CI acceptability with all explanatory variables, except for \textit{Trial order}, and confirms the existence of multicollinearity. Table \ref{tab:looic_opt_hack_new_data} supports the superiority of \textsc{mean} over \textsc{max} in predicting acceptability. Table \ref{tab:coefficients_opt_hack_new_data} even highlights the relative importance of \textsc{mean} where when the alternative \textsc{max} variable was entered as the single posterior variable, there was uncertainty of its importance such that the coefficient might contain zero, consistent with results in the exploration phase (Section \ref{sec:appendix_exploratory_phase}). Overall, results with OPT align with those with GPT-2 Small in supporting the superiority of the \textsc{mean} posterior model in predicting CI acceptability.

\begin{table}[htbp]
\centering
\begin{adjustbox}{width=0.7\linewidth}
\begin{tabular}{lrrrrrr}
\hline
\textbf{model} & \textbf{LOOIC} & \textbf{SE} & \textbf{$\Delta$ LOOIC} & \textbf{$\Delta$ SE}  \\ \hline
\textsc{max}, \textsc{mean} & $29400.48$ & $183.95$ & $0.00$ & $0.00$ \\
\textsc{mean}      & $29430.65$ & $183.86$ & $30.17$ & $-0.09$ \\
\textsc{max}       & $29527.99$ & $184.67$ & $127.51$ & $0.72$ \\
\textsc{base}      & $29528.59$ & $184.68$ & $128.01$ & $0.73$\\ \hline
\end{tabular}
\end{adjustbox}
\caption{\textbf{OPT}: Model comparison via LOOIC. (LOOIC = leave-one-out information criterion. Smaller LOOIC indicates better fit.)} \label{tab:looic_opt_hack_new_data}
\end{table}

\begin{table}[htbp]
\centering
\begin{adjustbox}{width=\linewidth}
\begin{tabular}{lccccc}
\hline
\textbf{model} & \textbf{SLOR} & \textbf{Trial Order} & \textbf{Baseline} &
\textbf{$f_{\text{max}}$} & \textbf{$f_{\text{mean}}$} \\ \hline
max, mean & $0.158\,[0.119,\,0.197]$ & $-0.070\;[-0.109,\,-0.031]$ & $0.140\,[0.101,\,0.179]$ & $-0.127\;[-0.171,\,-0.082]$ & $0.273\,[0.226,\,0.321]$ \\
mean      & $0.161\,[0.122,\,0.201]$ & $-0.073\;[-0.112,\,-0.035]$ & $0.140\,[0.102,\,0.179]$ & ---                       & $0.200\,[0.160,\,0.240]$ \\
max       & $0.211\,[0.174,\,0.249]$ & $-0.073\;[-0.112,\,-0.034]$ & $0.157\,[0.117,\,0.196]$ & $0.015\;[-0.023,\,0.053]$ & --- \\
base      & $0.213\,[0.175,\,0.252]$ & $-0.072\;[-0.110,\,-0.034]$ & $0.157\,[0.118,\,0.196]$ & ---                       & --- \\ \hline
\end{tabular}
\end{adjustbox}
\caption{\textbf{OPT}: Estimated coefficients of explanatory variables.
The estimate is the mean of the posterior distribution of the explanatory variable, with the 95\% Bayesian credible interval in square brackets.} \label{tab:coefficients_opt_hack_new_data}
\end{table}

\end{document}